\documentclass[letterpaper, 10 pt, conference]{IEEEtran}
\IEEEoverridecommandlockouts  

\usepackage{multicol}
\usepackage[bookmarks=true]{hyperref}

\usepackage{nicefrac}
\usepackage{mathtools}
\usepackage{amsthm}
\usepackage{amsmath}
\usepackage{paralist}
\usepackage[capitalise]{cleveref}

\usepackage[export]{adjustbox}
\usepackage{placeins}
\usepackage{makecell}

\theoremstyle{definition}

\newcommand{\framework}{\scenario{PerSyS}} 

\newcommand{\optional}[1]{{\color{gray}#1}}
\newcommand{\PMC}{PMC\xspace}

\newcommand{\maxFaults}{\kappa}
\newcommand{\myt}{t}

\newcommand{\tests}{tests\xspace}
\newcommand{\test}{test\xspace}

\newcommand{\ab}{_{[a,b]}}
 
\usepackage{comment}
\usepackage{siunitx}
\usepackage{relsize}
\usepackage{ifthen}
\usepackage[colorinlistoftodos]{todonotes}

\usepackage[caption=false]{subfig}

\usepackage[vlined,ruled,linesnumbered]{algorithm2e}
\usepackage{graphics} 
\usepackage{rotating}
\usepackage{color}
\usepackage{enumerate}
\usepackage[T1]{fontenc}
\usepackage{psfrag}
\usepackage{epsfig} 
\usepackage{booktabs}
\usepackage{graphicx,url}
\usepackage{multirow}
\usepackage{array}
\usepackage{latexsym}
\usepackage{amsfonts}
\usepackage{amsmath}
\usepackage{amssymb}
\usepackage{xstring}
\usepackage[noend]{algorithmic}
\usepackage{multirow}
\usepackage{xcolor}
\usepackage{prettyref}
\usepackage{flexisym}
\usepackage{bigdelim}
\usepackage{breqn} 
\usepackage{listings}
\let\labelindent\relax
\usepackage{enumitem}
\usepackage{xspace}
\usepackage{bm}
\graphicspath{{./figures/}}
\usepackage{tikz}
\usetikzlibrary{matrix,calc}

\usepackage{mdwlist}

\let\itemize\undefined
\makecompactlist{itemize}{stditemize}

\usepackage{amsthm}
\newtheoremstyle{mystyle}
  {}
  {}
  {\itshape}
  {}
  {\bfseries}
  {.}
  { }
  {\thmname{#1}\thmnumber{ #2}\thmnote{ (#3)}}
\theoremstyle{mystyle}

\newrefformat{prob}{Problem\,\ref{#1}}
\newrefformat{def}{Definition\,\ref{#1}}
\newrefformat{sec}{Section\,\ref{#1}}
\newrefformat{sub}{Section\,\ref{#1}}
\newrefformat{prop}{Proposition\,\ref{#1}}
\newrefformat{app}{Appendix\,\ref{#1}}
\newrefformat{alg}{Algorithm\,\ref{#1}}
\newrefformat{cor}{Corollary\,\ref{#1}}
\newrefformat{thm}{Theorem\,\ref{#1}}
\newrefformat{lem}{Lemma\,\ref{#1}}
\newrefformat{fig}{Fig.\,\ref{#1}}
\newrefformat{tab}{Table\,\ref{#1}}

\newtheorem{theorem}{Theorem}

\newtheorem{corollary}[theorem]{Corollary}

\newtheorem{definition}[theorem]{Definition}
\newtheorem{proposition}[theorem]{Proposition}

\newcommand{\bdmath}{\begin{dmath}}
\newcommand{\edmath}{\end{dmath}}
\newcommand{\beq}{\begin{equation}}
\newcommand{\eeq}{\end{equation}}
\newcommand{\bdm}{\begin{displaymath}}
\newcommand{\edm}{\end{displaymath}}
\newcommand{\bea}{\begin{eqnarray}}
\newcommand{\eea}{\end{eqnarray}}
\newcommand{\beal}{\beq \begin{array}{ll}}
\newcommand{\eeal}{\end{array} \eeq}
\newcommand{\beas}{\begin{eqnarray*}}
\newcommand{\eeas}{\end{eqnarray*}}
\newcommand{\ba}{\begin{array}}
\newcommand{\ea}{\end{array}}
\newcommand{\bit}{\begin{itemize}}
\newcommand{\eit}{\end{itemize}}
\newcommand{\ben}{\begin{enumerate}}
\newcommand{\een}{\end{enumerate}}

\newcommand{\calI}{{\cal I}}

\newcommand{\setE}{\textsf{E}}

\newcommand{\setU}{\textsf{U}}

\newcommand{\setal}{~\emph{et~al.}\xspace}
\newcommand{\eg}{\emph{e.g.,}\xspace}
\newcommand{\ie}{\emph{i.e.,}\xspace}
\newcommand{\myParagraph}[1]{{\bf #1.}\xspace}

\newcommand{\hide}[1]{}

\newcommand{\hiddenText}{{\color{gray} hidden text.}}
\newcommand{\hideWithText}[1]{\hiddenText}

\newcommand{\dist}{\mathbf{dist}}

\newcommand{\scenario}[1]{{\smaller \sf#1}\xspace}

\newcommand{\blue}[1]{{\color{blue}#1}}

\newcommand{\linkToPdf}[1]{\href{#1}{\blue{(pdf)}}}
\newcommand{\linkToPpt}[1]{\href{#1}{\blue{(ppt)}}}
\newcommand{\linkToCode}[1]{\href{#1}{\blue{(code)}}}
\newcommand{\linkToWeb}[1]{\href{#1}{\blue{(web)}}}
\newcommand{\linkToVideo}[1]{\href{#1}{\blue{(video)}}}
\newcommand{\linkToMedia}[1]{\href{#1}{\blue{(media)}}}
\newcommand{\award}[1]{\xspace} 

\newcommand{\rev}[1]{{#1}} 

\newcommand{\modules}{modules\xspace}
\newcommand{\module}{module\xspace}

\newcommand{\unit}{processor\xspace}
\newcommand{\units}{processors\xspace}

\newcommand{\uniti}{i\xspace}
\newcommand{\unitj}{j\xspace}

\newcommand{\obs}[1]{\mathcal{O}_\mathrm{#1}}
\newcommand{\approxin}{\>\widetilde{\in}\>}
\newcommand{\st}{\>\mbox{s.t.}\>}
\DeclareMathOperator{\fov}{FoV}

\DeclareSIUnit{\mph}{mph}

\newcommand{\ct}[2]{
  \vspace{-1mm}
  \begin{table}[htpb!]
  \begin{tabular}{ l l }
    \textbf{Edge} & {\hspace{0.5em}{#1}} \\
    \textbf{Check} & \adjustbox{valign=t}{ {#2} }
  \end{tabular}
  \end{table}
  \vspace{-3mm}
}

\newcommand{\toAdd}[1]{}

\tikzset{
graphnode/.style={
  circle,
  inner sep=0pt,
  text width=4mm,
  align=center,
  draw=black,
  fill=white
  }
}
\newcommand*\circled[1]{\tikz[baseline=(char.base)]{
            \node[graphnode] (char) {#1};}}
\newcommand*\node[1]{\circled{#1}}

\DeclarePairedDelimiter{\floor}{\lfloor}{\rfloor}

\newcommand{\apollo}{\scenario{Apollo Auto}}
\newcommand{\lgsvl}{\scenario{LGSVL}}

\newcommand{\mpPostSpace}{\vspace{-3mm}}
\newcommand{\mpColTwo}{9.15cm}
\newcommand{\mpMidSpaceTwo}{\hspace{-4mm}}

\newcommand{\mpColFour}{4.4cm}

\usepackage[prefix=tikzsym]{tikzsymbols}

\DeclarePairedDelimiter{\pair}{\langle}{\rangle}

\newcommand{\tn}{\textnormal}

\renewcommand{\ss}{\subseteq}

\newcommand{\from}{\leftarrow}

\usetikzlibrary{
	cd,
	math,
	decorations.markings,
	decorations.pathreplacing,
	decorations.pathmorphing,
	positioning,
	arrows.meta,
	circuits.logic.US,
	shapes,
	calc,
	fit,
	quotes}

\tikzset{
     oriented WD/.style={
        every to/.style={out=0,in=180,draw},
        label/.style={
           font=\everymath\expandafter{\the\everymath\scriptstyle},
           inner sep=0pt,
           node distance=2pt and -2pt},
        semithick,
        node distance=1 and 1,
        decoration={markings, mark=at position \stringdecpos with \stringdec},
        ar/.style={postaction={decorate}},
        execute at begin picture={\tikzset{
           x=\bbx, y=\bby,
           }}
        },
     string decoration/.store in=\stringdec,
     string decoration={\arrow{stealth};},
     string decoration pos/.store in=\stringdecpos,
     string decoration pos=.7,
     bbx/.store in=\bbx,
     bbx = 1.5cm,
     bby/.store in=\bby,
     bby = 1.5ex,
     bb port sep/.store in=\bbportsep,
     bb port sep=1.5,
     bb port length/.store in=\bbportlen,
     bb port length=4pt,
     bb penetrate/.store in=\bbpenetrate,
     bb penetrate=0,
     bb min width/.store in=\bbminwidth,
     bb min width=1cm,
     bb rounded corners/.store in=\bbcorners,
     bb rounded corners=2pt,
     bb small/.style={bb port sep=1, bb port length=2.5pt, bbx=.4cm, bb min width=.4cm, 
bby=.7ex},
		 bb medium/.style={bb port sep=1, bb port length=2.5pt, bbx=.4cm, bb min width=.4cm, 
bby=.9ex},
     bb/.code 2 args={
        \pgfmathsetlengthmacro{\bbheight}{\bbportsep * (max(#1,#2)+1) * \bby}
        \pgfkeysalso{draw,minimum height=\bbheight,minimum width=\bbminwidth,outer 
sep=0pt,
           rounded corners=\bbcorners,thick,
           prefix after command={\pgfextra{\let\fixname\tikzlastnode}},
           append after command={\pgfextra{\draw
              \ifnum #1=0{} \else foreach \i in {1,...,#1} {
                 ($(\fixname.north west)!{\i/(#1+1)}!(\fixname.south west)$) +(-
\bbportlen,0) 
  coordinate (\fixname_in\i) -- +(\bbpenetrate,0) coordinate (\fixname_in\i')}\fi 
              \ifnum #2=0{} \else foreach \i in {1,...,#2} {
                 ($(\fixname.north east)!{\i/(#2+1)}!(\fixname.south east)$) +(-
\bbpenetrate,0) 
  coordinate (\fixname_out\i') -- +(\bbportlen,0) coordinate (\fixname_out\i)}\fi;
           }}}
     },
     bb name/.style={append after command={\pgfextra{\node[anchor=north] at 
(\fixname.north) {#1};}}}
}

\tikzset{
  	unoriented WD/.style={
  		every to/.style={draw},
  		shorten <=-\penetration, shorten >=-\penetration,
  		label distance=-2pt,
  		thick,
  		node distance=\spacing,
  		execute at begin picture={\tikzset{
  			x=\spacing, y=\spacing}}
  		},
  	pack size/.store in=\psize,
  	pack size = 8pt,
  	spacing/.store in=\spacing,
  	spacing = 8pt,
  	link size/.store in=\lsize,
  	link size = 2pt,
		penetration/.store in=\penetration,
		penetration = 2pt,
  	pack color/.store in=\pcolor,
  	pack color = blue,
  	pack inside color/.store in=\picolor,
  	pack inside color=blue!20,
  	pack outside color/.store in=\pocolor,
  	pack outside color=blue!50!black,
  	surround sep/.store in=\ssep,
  	surround sep=8pt,
  	link/.style={
  		circle, 
  		draw=black, 
  		fill=black,
  		inner sep=0pt, 
  		minimum size=\lsize
  	},
  	pack/.style={
  		circle, 
  		draw = \pocolor, 
  		fill = \picolor,
  		inner sep = .25*\psize,
  		minimum size = \psize
  	},
  	outer pack/.style={
  		ellipse, 
  		draw,
  		inner sep=\ssep,
  		color=\pocolor,
  	},
  	intermediate pack/.style={
  		ellipse,
  		dashed, 
  		draw,
  		inner sep=\ssep,
  		color=\pocolor,
  	},
}

\graphicspath{figures}

\usepackage{float}
\usepackage{hyperref}
\usepackage{graphicx}
\usepackage{epstopdf}
\usepackage{epsfig}

\newcommand{\myparagraph}[1]{{\bf #1.}}

\graphicspath{{./figures/}}

\usepackage{xr}

\newcommand{\arxivVersion}[2]{{#1}\xspace} 

\newcommand{\isExtended}[2]{#1}
\newcommand{\suppMaterial}{Appendix\xspace}
\newcommand{\suppMaterialLong}{\suppMaterial\xspace}

\title{\vspace{18pt}\huge{Monitoring and Diagnosability of Perception Systems}\vspace{-10pt}}

\author{Pasquale Antonante, David I. Spivak, Luca Carlone
\thanks{P.\,Antonante, {D.\,I.\,Spivak}, and L.\,Carlone are with the Laboratory for 
Information \& Decision Systems, Massachusetts Institute of Technology, Cambridge, MA, USA.
{\sf \{antonap, dspivak, lcarlone\}@mit.edu}}
\thanks{This work was partially funded by the NSF CAREER award ``Certifiable Perception for Autonomous Cyber-Physical Systems'', AFOSR FA9550-20-1-0348 and by MathWorks. }
}

\begin{document}

\maketitle

\arxivVersion{
	\begin{tikzpicture}[overlay, remember picture]
	\path (current page.north east) ++(-1.8,-0.4) node[below left] {
	This paper has been published in the 2021 IEEE/RSJ International Conference on Intelligent Robots and Systems (IROS).
	};
	\end{tikzpicture}
	\begin{tikzpicture}[overlay, remember picture]
	\path (current page.north east) ++(-5.75,-0.8) node[below left] {
	Please cite the paper as: P. Antonante, D. I. Spivak, and L. Carlone,
	};
	\end{tikzpicture}
	\begin{tikzpicture}[overlay, remember picture]
	\path (current page.north east) ++(-0.5,-1.2) node[below left] {
	``Monitoring and Diagnosability of Perception Systems'', \emph{IEEE/RSJ International Conference on Intelligent Robots and Systems (IROS)}, 2021.
	};
	\end{tikzpicture}
}{}

\vspace{-1em}
\begin{abstract}
Perception is a critical component of high-integrity applications of robotics and autonomous systems, such 
as self-driving vehicles. 
 In these applications, failure of perception systems may put human life at risk, and 
 a broad adoption of these technologies requires the development of methodologies to guarantee and monitor
  safe operation.
Despite the paramount importance of perception systems, currently there is no formal 
approach for system-level monitoring. 
In this work, we propose a mathematical model for runtime monitoring and fault detection and identification in perception systems.
 Towards this goal, we draw connections with the literature on \emph{diagnosability} in multiprocessor systems, and 
generalize it 
to account for \modules with heterogeneous outputs that interact over time.
 The resulting \emph{temporal diagnostic graphs} 
 (i)~provide a 
 framework to reason over the consistency of perception outputs
  --across \modules and over time-- thus enabling fault detection, 
   (ii)~allow us to establish formal guarantees on the maximum number of faults that can be uniquely
  identified in a given perception system, 
  and (iii)~enable the design of efficient algorithms for fault identification. 
  We demonstrate our monitoring system, dubbed \emph{\framework}, in realistic simulations using the \lgsvl self-driving simulator 
  and the \apollo autonomy software stack,
  and show that \framework is able to detect failures in challenging scenarios (including scenarios that have caused self-driving car accidents in recent years), and 
   is able to correctly identify faults while entailing a minimal computation  overhead ($<\;$\SI{5}{\milli\second} on a single-core CPU).
\end{abstract}
\vspace{-0.5em}
\section*{Supplementary Material}

\begin{itemize}
	\item Video: {\footnotesize \url{https://youtu.be/GJ9esRS8jZs}}
\end{itemize}

\section{Introduction}\label{sec:introduction}

The automotive industry is undergoing a change that could revolutionize mobility.
Self-driving cars promise a deep transformation of personal mobility and have the potential to improve safety, 
efficiency (\eg commute time, fuel), and induce a paradigm shift in how entire cities are designed~\cite{Silberg12wp-selfDriving}.
One key factor that drives the adoption of such technology is the capability of ensuring 
and monitoring safe operation. 
Consider Uber's fatal self-driving crash~\cite{ntsbuber} in 2018: the report from the National Transportation Safety Board states that ``inadequate safety culture'' contributed to the fatal collision between the autonomous vehicle and the pedestrian. 
 The lack of safety guarantees, combined with the unavailability of formal monitoring tools, 
  is the root cause of these accidents and has a profound impact on the user's trust.
The American Automobile Association's survey~\cite{aaa} shows that 71\% of Americans claim to be afraid of riding in a self-driving car.
This is a clear sign that the industry needs a sound methodology, embedded in the design process, to guarantee 
safety and build public trust.


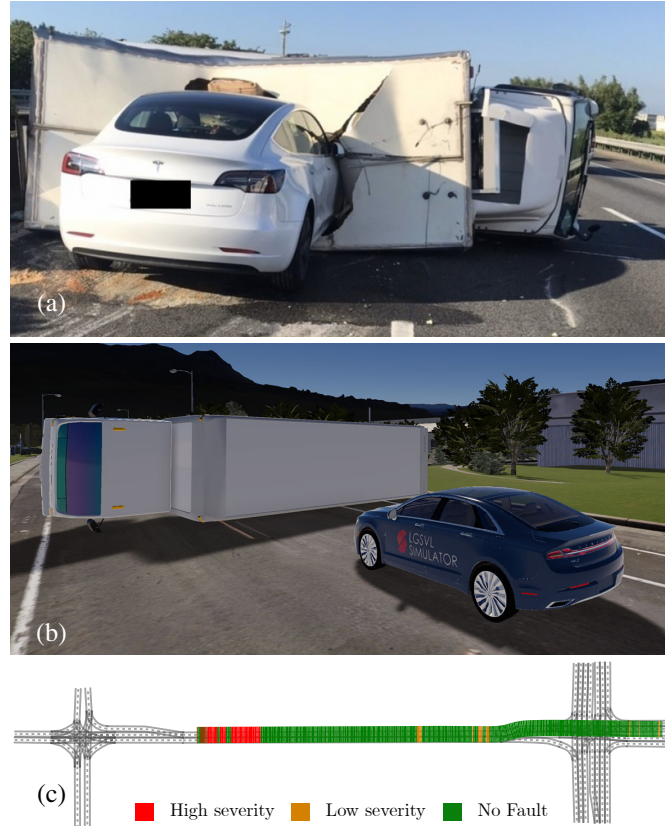
\begin{figure}[ht!]
  \setlength\belowcaptionskip{-10\baselineskip}
  \vspace{2.2mm}
	\begin{center}
  \begin{minipage}{\textwidth}
  \hspace{-18pt}
	\begin{tabular}{c}%
		\begin{minipage}{\mpColTwo}%
        \centering%
        \begin{tikzpicture}
        \draw (0, 0) node[inner sep=0] {\includegraphics[trim=0 50 0 20,clip,width=0.95\columnwidth]{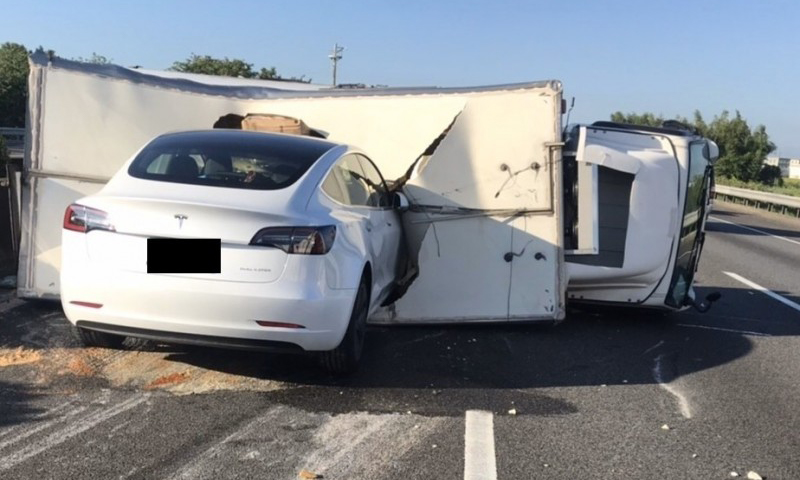}};
        \draw[text=white] (-3.8, -1.8) node {(a)};
        \end{tikzpicture}
		\end{minipage}
		\vspace{1mm}\\
		\begin{minipage}{\mpColTwo}%
        \centering%
        \begin{tikzpicture}
        \draw (0, 0) node[inner sep=0] {\includegraphics[trim=0 60 0 90,clip,width=0.95\columnwidth]{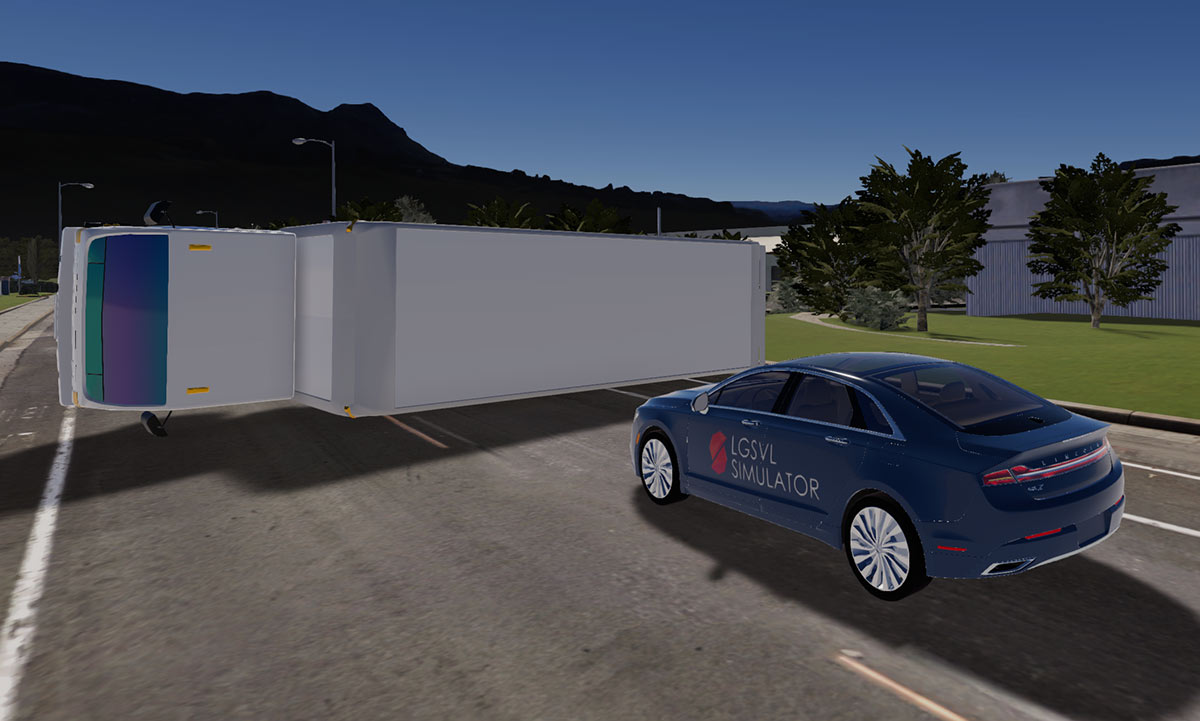}};
        \draw[text=white] (-3.8, -1.8) node {(b)};
        \end{tikzpicture}
		\end{minipage}
		\vspace{0.7mm}\\
		\begin{minipage}{\mpColTwo}%
        \centering%
        \begin{tikzpicture}
        \draw (0, 0) node[inner sep=0] {\includegraphics[width=0.95\columnwidth]{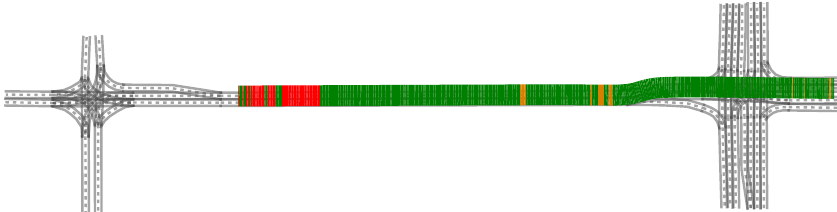}};
        \draw[text=black] (-3.8, -0.7) node {(c)};
        \end{tikzpicture}
    \end{minipage}
    \vspace{-4mm}\\
		\begin{minipage}{\mpColTwo}%
        \centering%
        \includegraphics[width=0.6\columnwidth]{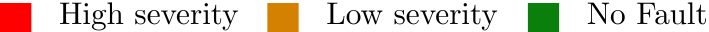} 
		\end{minipage}
		\end{tabular}
	\end{minipage}
  \caption{(a) Tesla Autopilot fails to detect an overturned truck, leading to a crash (Taiwan, 2020)~\cite{teslaAccidentTruck}. 
  			(b) We replicate the same scenario in the \lgsvl simulator~\cite{lgsvl-sim} and observe it still causes accidents in a state-of-the-art open-source autonomy stack (\apollo~\cite{apollo-auto}).
  			(c) We propose \emph{\framework}, a graph-theoretic framework for fault detection and identification, and show it 
  			is able to detect failures and prevent crashes in realistic simulations, including the scenario in (b).
  			 The colors in the figure show the result of the fault detection.
        \label{fig:front_page}}
  \end{center}
  \vspace{-12mm}
\end{figure}

While safety guarantees have been investigated in the context of control and decision-making~\cite{Mitsch17ijrr-verificationObstacleAvoidance,Foughali18formalise-verificationRobots}, 
the state of the art is still lacking a formal and broadly applicable methodology for monitoring \emph{perception systems}, which 
constitute a key component of any autonomous vehicle. Perception systems provide functionalities such as localization and obstacle mapping, lane detection, detection and tracking of other vehicles, pedestrians, and traffic signs, among others.

\myParagraph{State of Practice}
The automotive industry currently uses 
four classes of methods to claim the safety of an autonomous vehicle (AV)~\cite{Shalev-Shwartz17arxiv-safeDriving}, namely: miles driven, simulation, scenario-based testing, and disengagement. 
Each of these methods has well-known limitations. 
The \emph{miles driven} approach is based on the statistical argument that if the probability of crashes per mile is lower in autonomous vehicles than for humans, then AVs are safer; however, such an analysis would 
 require an impractical amount (\ie\!billions) of miles to produce statistically-significant results~\cite{Kalra16tra-selfDriving,Shalev-Shwartz17arxiv-safeDriving}.\footnote{Moreover, 
the analysis should cover all representative driving conditions (\eg driving on a highway is easier than driving 
in urban environment) and should be repeated at every software update, quickly becoming impractical.} 
 The same approach can be made more scalable through \emph{simulation}, but unfortunately 
  creating a life-like simulator is an open problem, for some aspects even more challenging than self-driving itself. 
 \emph{Scenario-based} testing is based on the idea
 that if we can enumerate all the possible driving scenarios that could occur, then we can simply expose the AV (via simulation, 
 closed-track testing, or on-road testing) to all these scenarios and, as a result, be confident that the AV will only make sound 
  decisions. 
However, enumerating all possible corner cases (and perceptual conditions) is a daunting task. 
Finally, \emph{disengagement}~\cite{googledisengagements} is defined as the moment when a human safety driver has to intervene in order to prevent a hazardous situation. 
However, while less frequent disengagements indicate an improvement of the AV behavior, they do not give evidence of the system safety.

An established methodology to ensure safety 
is to develop a \emph{standard} that every manufacturer has to comply with.
In the automotive industry, the standard ISO 26262~\cite{iso26262} 
is a risk-based safety standard that applies to electronic systems in production vehicles.
A key issue is that ISO 26262 relies on the presence of human drivers (and mostly focuses on electronic systems rather than algorithmic aspects) hence it does not readily apply to fully autonomous vehicles~\cite{Koopman16sae-autonomousVehicleTesting}.
The recent ISO 21448~\cite{iso21448}  
is intended to complement ISO 26262, 
but currently mostly provides high-level considerations to inform stakeholders.
The report~\cite{AptivSafetyAV} discusses verification and validation for highly autonomous vehicles, 
and 
 suggests the use of \emph{monitors} to detect 
off-nominal performance. 
Perception has been the subject to increasing attention also  
outside the automated driving industry~\cite{UberElevate, EHangAirMobility}.
The report~\cite{EASADesignNN} considers an automatic aircraft landing system as case study and investigates the challenges in developing trustworthy AI, with focus on machine learning. 
These reports stress the role of perception for autonomous systems and 
motivate us to design a rigorous framework for system-level monitoring.

\myParagraph{Related Work} 
Formal methods~\cite{Ingrand19irc-verificationTrends,Seshia16arxiv-verifiedAutonomy,Ding14icra-hierarchicalPlanning,Desai17icrv-verification,Hoxha16aaaiws-temporalLogic,Vasile17icra-motionPlanning, Dathathri17arxiv-synthesis,Ghosh16hscc-formalMethods,Li11fmmc-formalMethods,Li14tacas-formalMethods,Kloetzer08tac-temporalLogic,Roohi18arxiv-selfDrivingVerificationBenchmark,Mitsch17ijrr-verificationObstacleAvoidance}
have been recently used as a tool to study safety of autonomous systems.
These approaches have been successful for decision systems, such as obstacle avoidance~\cite{Mitsch17ijrr-verificationObstacleAvoidance}, road rule compliance~\cite{Roohi18arxiv-selfDrivingVerificationBenchmark} and high-level decision-making~\cite{Cardoso2020nasa-curiosityVerification}, where  
the specifications are usually model-based and have well-defined semantics~\cite{Foughali18formalise-verificationRobots}.
However, they are challenging to apply to perception systems, due to the complexity of modeling the physical environment~\cite{Seshia16arxiv-verifiedAutonomy}, and the trade-off between evidence for certification and tractability of the model~\cite{Luckcuck19csur-surveyFormalMethods}.
Current approaches~\cite{Dreossi19arxiv-verifai,Fremont19ieee-scenic,Jha17ar-SafeAutonomy,Leahy19ijrr-formalMethods} consider high-level abstractions of perception~\cite{Shalev-Shwartz17arxiv-safeDriving,Balakrishnan19date-perceptionVerification,Dokhanchi18rv-perceptionVerification} or rely on simulation to assert the true state of the world~\cite{Dreossi19arxiv-verifai,Fremont19ieee-scenic,Dreossi17arxiv-testingDNN}. 

Relevant to the approach presented is this paper is the class of \emph{runtime verification} methods.
Runtime verification is an online approach based on extracting information from a running system and using it to detect (and possibly react to) observed behaviors satisfying or violating certain properties~\cite{Bartocci18book-runtimeVerification, Francalanza18book-runtimeVerification}.
Traditionally, the task of evaluating whether a \module is working properly is assigned to a monitor, which verifies some input/output properties on the \module alone.
Balakrishnan\setal~\cite{Balakrishnan19date-perceptionVerification} use Timed Quality Temporal Logic (TQTL) to reason about desiderable spatio-temporal properties of a perception algorithm.
Kang\setal~\cite{Kang18nips-ModelAF} use model assertions, which similarly place a logical constraint on the output of a \module to detect anomalies.
We argue that this paradigm does not fully capture the complexity of perception pipelines:
while monitors can be used to infer the state of a \module, perception pipelines provide an additional opportunity to cross-check the \emph{compatibility} of results across different \modules. 
In this paper we try to address this limitation.

Fault-tolerant architectures~\cite{navarro20arxiv-hero} have also been proposed to detect and potentially recover from a faulty state, but these efforts mostly focus on implementing watchdogs and monitors for specific modules, rather than providing tools for system-level analysis and monitoring.

Finally, performance guarantees for perception have been investigated for specific problems. 
In particular, related work on \emph{certifiable algorithms}~\cite{Yang19rss-teaser, Yang20cvpr-shapeStar, Briales18cvpr-global2view} provides algorithms that are capable of identifying faulty behaviors during execution. 
These related works mostly focus on specific algorithms, while our goal is to 
establish monitoring for perception \emph{systems}, including multiple interacting modules and algorithms.

\myParagraph{Contribution}
In this paper we develop a methodology to detect and identify faulty \modules in a perception pipeline at runtime.
In particular, we address two questions
(adapted from~\cite{Brundage20arxiv-trustworthyAI}):
\begin{inparaenum}[(i)] 
  \item \emph{Can I (as a developer) verify that the perception \modules are providing reliable interpretations of the sensor data?}
  \item \emph{Can I (as regulator) trace the steps that led to an accident caused by an autonomous vehicle?}
\end{inparaenum}
Towards this goal, our first contribution is to
 draw connections between perception monitoring and the literature on diagnosability in multiprocessor systems, and in particular 
 the \PMC 
 model~\cite{Preparata67tec-diagnosability}.
 Our second contribution is to generalize the \PMC model  to 
  account for \modules with heterogeneous outputs, and add a temporal dimension to the problem to account for \modules interacting over time. This results in the notion of \emph{temporal diagnostic graphs}, which
 (i) provide a 
 framework to reason over the consistency of perception outputs
  --across \modules and over time-- thus enabling failure detection, 
   (ii) allow us to establish formal guarantees on the maximum number of faults that can be uniquely
  identified in a given perception systems,\optional{\footnote{Our notion of \emph{diagnosability} is related to the level of redundancy within the system and provides a quantitative measure of robustness.}} 
  and (iii) enable the design of efficient algorithms for fault identification. 
  Our third contribution is to demonstrate our monitoring system, dubbed \framework, 
  in realistic simulations using the \lgsvl self-driving simulator 
  and the \apollo autonomy software stack,
  and show that \framework is able to detect failures in challenging scenarios (including night and adverse weather),
   including conditions that have caused self-driving car accidents in recent years (Fig.~\ref{fig:front_page}).
   Moreover, we show that \framework
   is able to correctly identify faults --whenever the system is sufficiently \emph{diagnosable}--
    while entailing a minimal computation  overhead ($<\;$\SI{5}{\milli\second} on a single-core CPU).

\section{Monitoring of Perception Systems}\label{sec:monitoring}

This section presents a mathematical model for monitoring and fault diagnosis in perception systems.
 Our model \emph{detects} if one of the  \modules in a perception system is returning an incorrect
 output, and possibly \emph{identifies} the faulty modules.
Section~\ref{sec:diagnosability} reviews 
results on fault detection and \emph{diagnosability} 
in multiprocessors systems, which has been extensively studied since the late 1960s.
Sections \ref{sec:tdg} and~\ref{sec:diagnosabilityPerception} extend diagnosability to perception systems and discuss how to 
model temporal aspects arising in AV applications.

\subsection{Diagnostic Graphs and Diagnosability}\label{sec:diagnosability}
Our approach builds on the \PMC model~\cite{Preparata67tec-diagnosability} for fault 
diagnosis.
In \PMC, 
a set of \emph{processors} 
is assembled such that each processor has the capability to communicate with a subset of the other processors, and all of the processors perform the same computation.
In such system, a fault occurs whenever the outputs of some processors disagree with each other; 
the problem is then to identify which \unit is faulty.
In the PCM model, each processor is assigned a subset of the other processors for the purpose of testing.
Using a comparison-based mechanism, the model aims to characterize the set of faulty processors.
Clearly, it is not possible to determine the faulty subset in general, so much of the literature on multiprocessor diagnosis considers two fundamental questions~\cite{Dahbura84tc-diagnosability}:
\begin{inparaenum}[(i)] 
  \item Given a collection of \units and a set of tests, what is the maximum number of arbitrary \units that can be faulty such that the set of faulty \units can be uniquely identified?
  \item Given a set of test results, does there exist an efficient procedure to identify the faulty \units?
\end{inparaenum}
The key tool to address these questions is the \emph{diagnostic graph}~\cite{Preparata67tec-diagnosability}.

\myParagraph{Diagnostic Graph}
At any given time, each \unit is assumed to be in one of two states: \emph{faulty} or \emph{fault-free}. 
Diagnosis is based on the ability of \units to test---\ie to check the consistency of its output against--- other \units.
Formally we assume that each \unit implements one or more \emph{consistency functions}; these are Boolean functions that return \emph{pass} ($0$) or \emph{fail} ($1$), depending on whether the output of two \units is in agreement. 

To perform the diagnosis, we follow~\cite{Preparata67tec-diagnosability} and model the problem as a directed graph $D=(U, E)$, where $ U $ is the set of \units, 
while the edges $E$ represent the test assignments.
In particular, for an edge $(\uniti,\unitj)\in E$, we say that node $\uniti$ is testing node $\unitj$.
The outcome of this test is the result of the consistency functions of $\uniti$, in other words $1$ (resp. $0$) if $\uniti$ evaluates $\unitj$ as faulty (resp. fault-free).
Fault-free \units are assumed to provide correct test results, whereas no assumption is made about tests executed by faulty \units: they may produce correct or incorrect test outcomes. 
We call $D$ the \emph{diagnostic graph}. 
The collection of all test results for a test assignment $E$ is called a \emph{syndrome}.
Formally, a syndrome is a function $\sigma:E\to\{0, 1\}$.
The syndrome is then processed to 
make some determination about the faultiness status of every \unit in the system.
The notion of $\maxFaults$-diagnosability formalizes when it is possible to use the syndrome to diagnose and identify all faults in a system.

\begin{definition}[$\maxFaults$-diagnosability~\cite{Preparata67tec-diagnosability}]\label{def.t_diag}
  A diagnostic graph $D=(U, E)$ with $|U|$ processors is \emph{$\maxFaults$-diagnosable} (with $\maxFaults< |U|$) if, given any syndrome, all faulty \units can be identified, provided that the number of faults does not exceed $\maxFaults$.
\end{definition}

The problem of determining the maximum value of $\maxFaults$ for which a given system is $\maxFaults$-diagnosable is called the \emph{diagnosability problem}.
We denote the maximum value of $\maxFaults$-diagnosability of a graph $D$ by $\maxFaults(D)$.

\myparagraph{Characterization of $\maxFaults$-diagnosability}
Consider a directed graph $D=(U ,E)$. 
For $\uniti \in U $, we denote with $\delta_{\tn{in}}(\uniti)$ the number of edges directed toward $\uniti$ (in-degree). We denote by $\delta_{\tn{in}}(D) = \min_{\uniti\in U } d_{\tn{in}}(\uniti)$ the minimum in-degree of the graph.
We denote with $\Gamma(\uniti) = \{\unitj \in U \mid (\uniti,\unitj) \in E\}$ the \emph{testable} set of $\uniti$ (outgoing neighbors of $\uniti$).
Finally, for $ X\subset U$, we define $\Gamma(X) = \{\bigcup_{\uniti \in X} \Gamma(\uniti) \setminus X\}$.
We can now state the following theorem that shows how to check if a graph is $\maxFaults$-diagnosable.
\begin{theorem}[Characterization of $\maxFaults$-diagnosability~\cite{Hakimi74tc-diagnosability}]\label{thm:pcm}
Let $D=( U ,E)$ be a diagnostic graph with $| U |$ \units. Then $D$ is $\maxFaults$-diagnosable if and only if
\begin{enumerate}[label=(\roman*)]
    \item\label{itm:t_ub} $\maxFaults \leq \frac{| U |-1}{2}$;
    \item\label{itm:t_lb} $\maxFaults \leq \delta_{\tn{in}}(D)$; and
    \item\label{itm:t_p} for each integer $p$ with $0\leq p<\maxFaults$, and each $ X\subset U $ with $| X|= |U|-2\maxFaults+p$ we have $|\Gamma( X)| > p$.
\end{enumerate}
\end{theorem}

A naive implementation of the checks in \cref{thm:pcm} would lead to an algorithm of time complexity $O(|U|^{t+2})$~\cite{Bhat82acm-diagnosability}.
Bhat~\cite{Bhat82acm-diagnosability} proposes an improved algorithm to find the maximum $\maxFaults$ of an arbitrary graph in $O(|U|^{u+2} \log_2(u))$ where $u=\min \{ \delta_{\tn{in}}(D), \floor{(|U|-1)/2}\}$.
Since this approach can be impractical for large graphs, they also propose a polynomial-time algorithm to find a lower-bound for $\maxFaults$ in $O(|U|^{3/2} +|E|)$.

\myparagraph{Fault Identification}
Once we have a syndrome on a $\maxFaults$-diagnosable graph, we would like to actually identify the set of faulty \units, provided that the number of faults does not exceed $\maxFaults$.
Dahbura\setal~\cite{Dahbura84tc-diagnosability} show that the problem of identifying the set of faulty \units is related to the problem of finding the minimum vertex cover set of an undirected graph. 
In general, the problem of finding a minimum vertex cover is in the class of NP-complete problems, meaning that there is no known deterministic algorithm that is guaranteed to solve the problem in polynomial time, but the validity of any solution can be tested in polynomial time.
However, the work~\cite{Dahbura84tc-diagnosability} exploits some special properties of $\maxFaults$-diagnosable graphs to propose an algorithm with time complexity $O(|U|^{2.5})$ for fault identification.
Sullivan~\cite{Sullivan88tc-diagnosability} proposes an algorithm with time complexity $O(\maxFaults^3+|E|)$ for fault identification and proves this is the tightest bound if $\maxFaults$ is $o(|U|^{5/6})$.
These results reassure us that the framework of diagnostic graphs provides a practical basis to build 
our real-time perception system monitor.

\subsection{Temporal Diagnostic Graphs}
\label{sec:tdg}

The \PMC model focuses on instantaneous fault diagnosis. 
In other words, given a diagnostic graph and a syndrome at time~$\myt$, 
\PMC-based algorithms detect and possibly identify faults only considering tests occurring at that time instant.
However, perception \modules evolve over time, are asynchronous, and output data at different rates.  
In the following, we extend the notion of diagnostic graphs to account for the temporal dimension of perception. 
Considering the temporal dimension will allow us 
to check consistency over time (we describe the temporal consistency \tests in Section~\ref{sec:diagnosabilityPerception}). 


Intuitively, if we consider an arbitrary time interval $[a,b]$, over the interval we have the opportunity to collect 
 multiple outputs for each \module in a perception system. 
Furthermore, the outputs of these \modules cannot be arbitrary, and must have some temporal consistency.
For instance, because the vehicle is moving at a finite velocity, measurements from the GPS at consecutive times should be within a maximum distance.
Similarly, since external objects detected by the perception system are stationary or move at bounded speed, consecutive detections must not be too spaced apart.
Therefore, we can build a diagnostic graph over the interval $[a,b]$ and test consistency of the perception results across \modules and over 
time, and detect failures when these \tests fail.

More formally, we define  an {\bf $[a,b]$-temporal diagnostic graph} $D\ab$ as a directed graph where each node 
represents the execution of a \module in the system within the time interval $[a,b]$. Each \module (\eg sensor fusion, object detection, 
LIDAR sensor) 
can be
executed multiple times in $[a,b]$, hence inducing multiple nodes in $D\ab$. 
Moreover, the graph has edges whenever an arbitrary pair of \modules can test (instantaneously or over time) 
their inputs or outputs. 
Intuitively, one can think at a temporal diagnostic graph as a larger graph where we ``stack'' 
multiple diagnostic graphs for each time $\myt \in [a,b]$, and add temporal \tests. 

Besides being more expressive, 
temporal diagnostic graph can improve diagnosability.
We formalize this aspect by first observing that adding edges (\eg temporal \tests) in the graph can only improve diagnosability%
\isExtended{
  (all proofs in~\suppMaterial).
}{
  (all proofs are given in~\suppMaterial).
}

\begin{corollary}\label{cor:more_edges}
If $D\ss D'$ are two graphs with the same set of vertices, then $\maxFaults(D)\leq \maxFaults(D')$.
\end{corollary}

\renewcommand{\setU}{U}
\renewcommand{\setE}{E}

Then we observe that connecting multiple $\maxFaults$-diagnosable graphs over time cannot decrease the diagnosability.
\begin{proposition}\label{prop:subgraph_diag}
Let $D\!=\!(\setU,\setE)$ be a diagnostic graph. 
Given subsets $\setU_i\ss\setU$ that cover $\setU$ (in the sense that $\setU=\cup_{i\in I}U_i$), and for each $i\in I$, let $D_i\ss D$ be the largest subgraph with vertices $\setU_i$.
Then if each $D_i$ is $\maxFaults$-diagnosable, so is $D$.
\end{proposition}

In the experimental section, we will observe that if the temporal \tests are well-designed, the 
diagnosability strictly increases when considering larger time intervals.

\subsection{Consistency Functions in Perception Systems}\label{sec:diagnosabilityPerception}

Each edges in our temporal diagnostic graph implements
a consistency function to \test consistency between the processing at different \modules in the perception system. 
This section defines four classes of consistency \tests for a perception system with arbitrary \modules interacting over time.

\subsubsection{Input Admissibility}
In this type of consistency \test, 
a \module $i$ monitors its inputs $\calI$ to judge whether they are admissible (\eg the IMU measurements fall within an admissible range).
This in turns induces a consistency \test between \module $i$ and another module $j$ producing $\calI$. 
Admissible ranges are known a priori in many perception \modules.

\subsubsection{Input or Output Consistency}
In this type of \test,  a \module monitors the consistency of inputs coming from (or outputs produced by) multiple other \modules.
For instance, consider two \modules $i$ and $j$ performing object detection using two different sensors, \eg camera and LIDAR.
Both \modules output a list of obstacles and, by comparison, $i$ can verify that $j$ is producing a consistent detection.
This class of \tests is common in perception, due to sensor redundancy.

\subsubsection{Input/Output Consistency}
In this class we include all \tests implemented by sensor fusion \modules, which \test their fused estimate against the input data.
Input/Output Consistency differs from Input Consistency in that it needs to compute the output variable in order to test the validity of the inputs.
As an example, consider the case in which Visual Odometry (\module $j$) incorrectly estimates the motion of the vehicle.
After fusing multiple incoming odometry measurements, the sensor fusion \module $i$ can identify that visual odometry 
produced an incorrect measurement~\cite{Lajoie19ral-DCGM,Antonante21tro-outlierRobustEstimation,Yang20neurips-certifiablePerception}. 

\subsubsection{Temporal Consistency}
In this class we include all the \tests between a \module $i$ and a \module $j$ involving data produced at different timestamps.
The simplest form of \emph{temporal consistency} involves an edge connecting two nodes corresponding to the outputs of the 
same \module at consecutive time steps, \eg $i$ and $j$ being two consecutive IMU measurements or pose estimates. 
The \tests can be extended to involve different \modules and non-consecutive timestamps.

\rev{
\subsection{Discussion}\label{sec:discussionConsistencyFunctions}
The diagnostic graph is a \emph{directed graph}. 
An edge $(i,j)$ indicates that \module $i$ tests $j$, reporting if the output of $j$ is correct from the standpoint if $i$.
The direction of the edge is useful whenever the test is non-symmetric, for example when a \module (\eg a watchdog) is testing the output of another \module (\eg state estimation) but not vice-versa.
When the test is symmetric (\ie two \modules test if their outputs are compatible with each other), 
both edge directions are theoretically valid (in our tests, we assign an arbitrary direction to the edge).
The binary outcome of a consistency \test asserts consistency across the \modules' outputs.

The perception system experiences a fault whenever two modules produce an inconsistent representation of the environment.
The faults can be intermittent, that is, occur at one time and disappear at the next.
Moreover, a consistency \test can implement an arbitrary logic to detect an inconsistency as long as it returns a boolean value indicating whether it detected an inconsistency (fault) or not. 
For instance, 
consider an object detection system including camera and LIDAR, where each detected object has a confidence level associated with it.
The consistency \test is tasked with determining if two sets of obstacles detected by the two sensors are in agreement.
The LIDAR might detect an object on the left of the car while the front-facing camera fails to do so.
This particular case may not be signaled as inconsistency if the object seen by the LIDAR is outside the camera's 
field of view.
Similarly, if the camera detects an obstacle with low confidence, while the LIDAR does not detect it, the consistency \test might be setup not to signal the inconsistency as the confidence is low.

The use of consistency functions generalizes the concept of \emph{voting} in estimation (\eg~\cite{Yang20tro-teaser}), enabling heterogeneous tests and arbitrarily complex logics to detect inconsistencies.
For example, while a consistency function verifying that a bounding box associated with a pedestrian occupies a meaningful region of the map, \ie sidewalks instead of the middle of a highway, can be easily incorporated into our system, 
it may be hard to include in standard estimation frameworks.

Finally, we note that our framework can also deal with the case where a \module produces a correct output despite using an incorrect input. 
For instance, several robust estimation schemes (\eg~\cite{Antonante21tro-outlierRobustEstimation}) can produce accurate results despite being fed with incorrect measurements.
}

\section{Experimental evaluation}

This section shows that temporal diagnostic graphs are effective to detect and identify failures in 
perception systems in real-time. 
We validate our theory in both numerical tests (Section~\ref{sec:exp-randomizedGraph}) 
and self-driving car simulations (Section~\ref{sec:exp-AV}). 

\myparagraph{Implementation Details} 
We build temporal diagnostic graphs in Python and perform fault identification using a constraint programming implementation of~\cite{Sullivan88tc-diagnosability,Dahbura84tc-diagnosability} using Google OR-Tools~\cite{google-ortools}.
\rev{The code is executed on a Linux machine with an Intel i-97920X processor (4.3 GHz) and two NVIDIA GeForce GTX 1080 Ti graphics cards.}
We call the resulting system \emph{\framework} (PERception SYstem Supervisor).
The specific choice of consistency \tests is discussed   below.

\subsection{Fault Identification in Diagnostic Graphs}
\label{sec:exp-randomizedGraph}

\begin{figure*}[ht!]
	\begin{center}
	\begin{minipage}{0.88\textwidth}
	\begin{tabular}{ccc}%
		\hspace{-14mm}
		\begin{minipage}{\mpColTwo}%
        \centering%
        \begin{tikzpicture}
        \draw (0, 0) node[inner sep=0] {\includegraphics[width=\columnwidth]{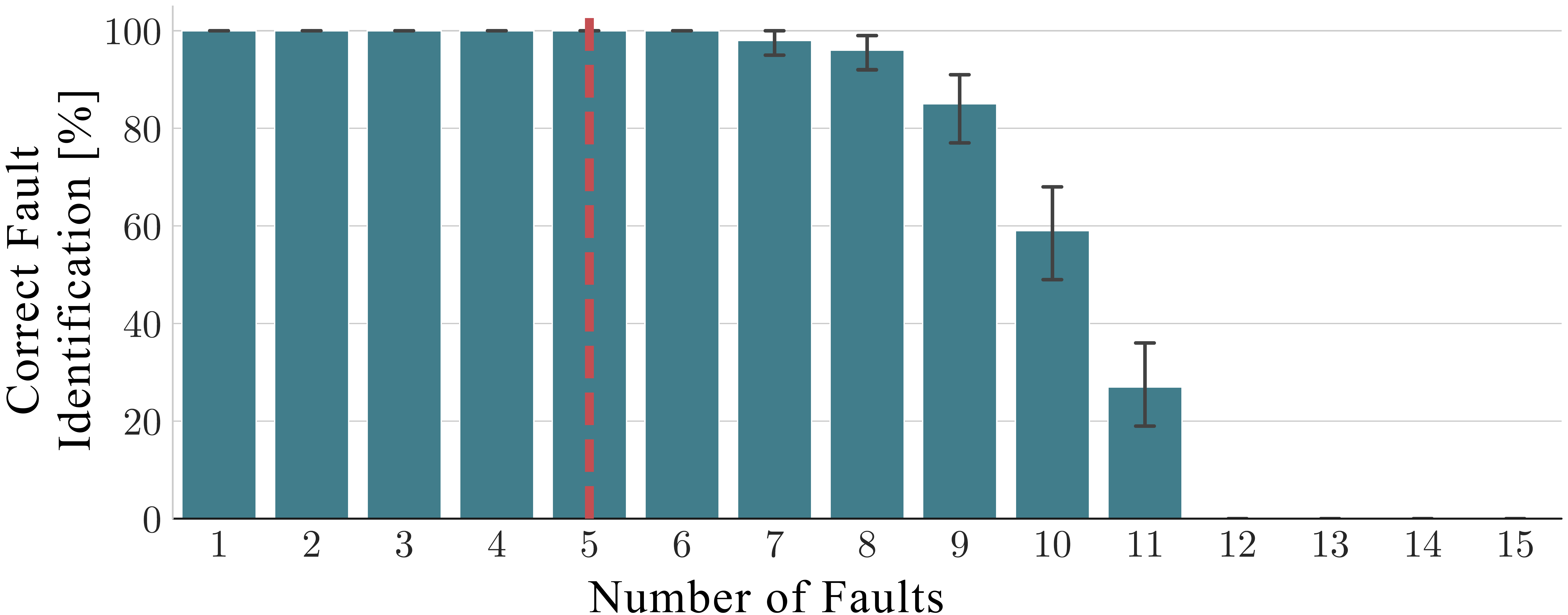}};
        \draw[text=black] (-4, -1.7) node {\footnotesize(a)};
        \end{tikzpicture}
			\end{minipage}
		& \mpMidSpaceTwo
			\begin{minipage}{\mpColTwo}%
        \centering%
        \begin{tikzpicture}
          \draw (0, 0) node[inner sep=0] {\includegraphics[width=\columnwidth]{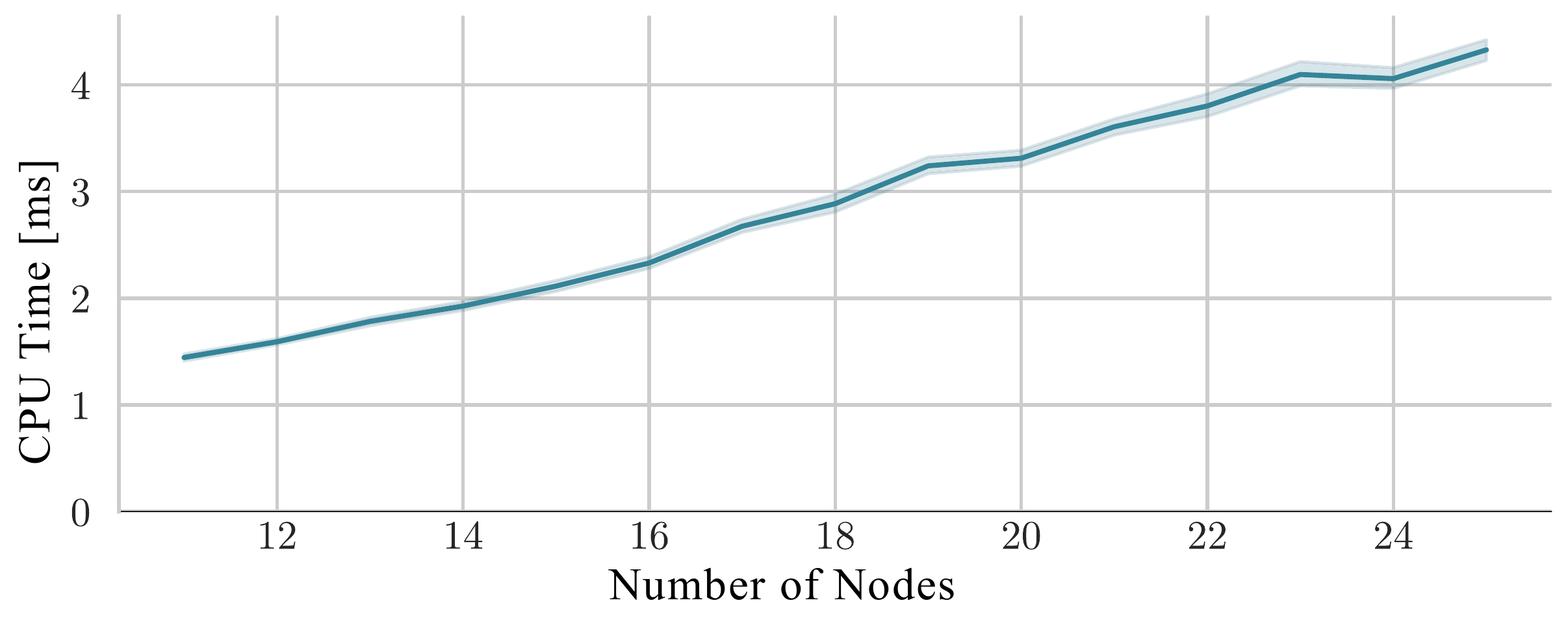}};
          \draw[text=black] (-4, -1.6) node {\footnotesize(b)};
          \end{tikzpicture}
			\end{minipage}
		\end{tabular}
	\end{minipage}
	\vspace{-2mm}
  \caption{ {\bf Diagnosability and fault identification in temporal diagnostic graphs.} (a) Percentage of correctly identified faults for randomly generated $5$-diagnosable graphs with $15$ nodes; (b) CPU time with 5 faults and increasing number of nodes.
  The plots also show the 1-$\sigma$ standard deviation.}
	\label{fig:toy_graph}
	\vspace{-5mm} 
	\end{center}
\end{figure*}

We start by evaluating \framework's capability of identifying faults. 
We perform a Monte Carlo analysis over randomly generated temporal diagnostic graphs.
To test the accuracy of the fault identification, measured as the percentage of correctly identified faults, we generate random $5$-diagnosable graphs with $15$ nodes and collect the fault identification results for increasing number of faults.
For each tested number of faults, we average the results over $100$ random graph.
\cref{fig:toy_graph}a shows that \framework is able to correctly identify all the faults 
whenever their number is below the theoretical limit of $5$ (remember that we simulate $5$-diagnosable graphs); 
in the presence of more than 5 faults, \framework is still able to \emph{detect} a failure, but its capability 
of \emph{identifying} which \modules are faulty decreases for increasing amounts of faults.
 \cref{fig:toy_graph}b considers a $5$-diagnosable graphs with $5$ randomly generated faults  
 and evaluates the CPU time for increasing number of nodes (from \num{11} to \num{25}).
The time required to identify the faults grows linearly with the number of nodes and does not exceed \SI{5}{\milli\second} for graphs with up to \num{25} nodes.


\subsection{Fault Detection in AV Perception Systems}
\label{sec:exp-AV}

We use \framework to monitor a state-of-the-art perception system, implemented in the open-source Baidu's \apollo~\cite{apollo-auto} autonomous driving platform.
We test the system \rev{on the Lincoln MKZ car, simulated} using LG's \lgsvl~\cite{lgsvl-sim} driving simulator. 
\rev{In our experiments we run on the same Linux machine the simulator, the AV software stack, and \framework in real-time.}
We test \framework in object detection and vehicle localization problems and show it 
is able to prevent accidents in the majority of the tests, even in adverse weather conditions.
\rev{Following a functional system architectures~\cite{Tacs2016iv-functionalArchitectureAV}, which separates the perception system from the decision system, when \framework identifies a fault in the perception system, it notifies the decision system that is in charge of reacting to it.}
\subsubsection{3D Object Detection}
We use \framework to detect failures in the subsystem in charge of 
detecting and localizing objects around the AV (\eg other vehicles, pedestrians).

\begin{figure}[t]
  \vspace{2mm} 
  \begin{center}
  \hspace{-3mm}
  \begin{minipage}{0.95\columnwidth}
  	\begin{tabular}{cc}%
  		\hspace{-4mm}
  		\begin{minipage}{\mpColFour}%
        \centering%
        \begin{tikzpicture}
        \draw (0, 0) node[inner sep=0] {\includegraphics[width=\columnwidth]{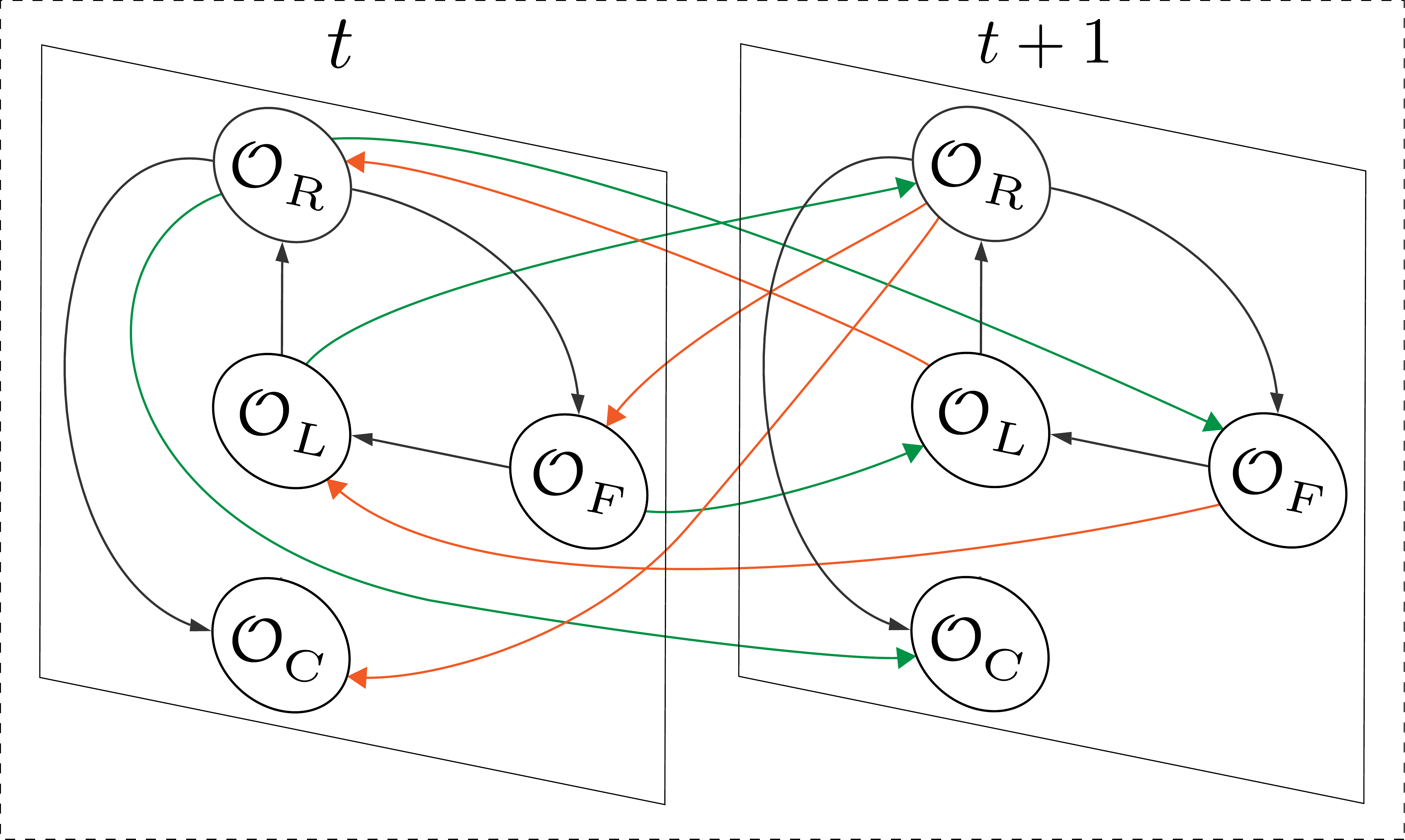}};
        \draw[text=black] (-1.85, -1.1) node {\footnotesize(a)};
        \end{tikzpicture}
  		\end{minipage}
  		& \hspace{-4mm}
  		\begin{minipage}{\mpColFour}%
        \centering%
        \begin{tikzpicture}
        \draw (0, 0) node[inner sep=0] {\includegraphics[width=\columnwidth]{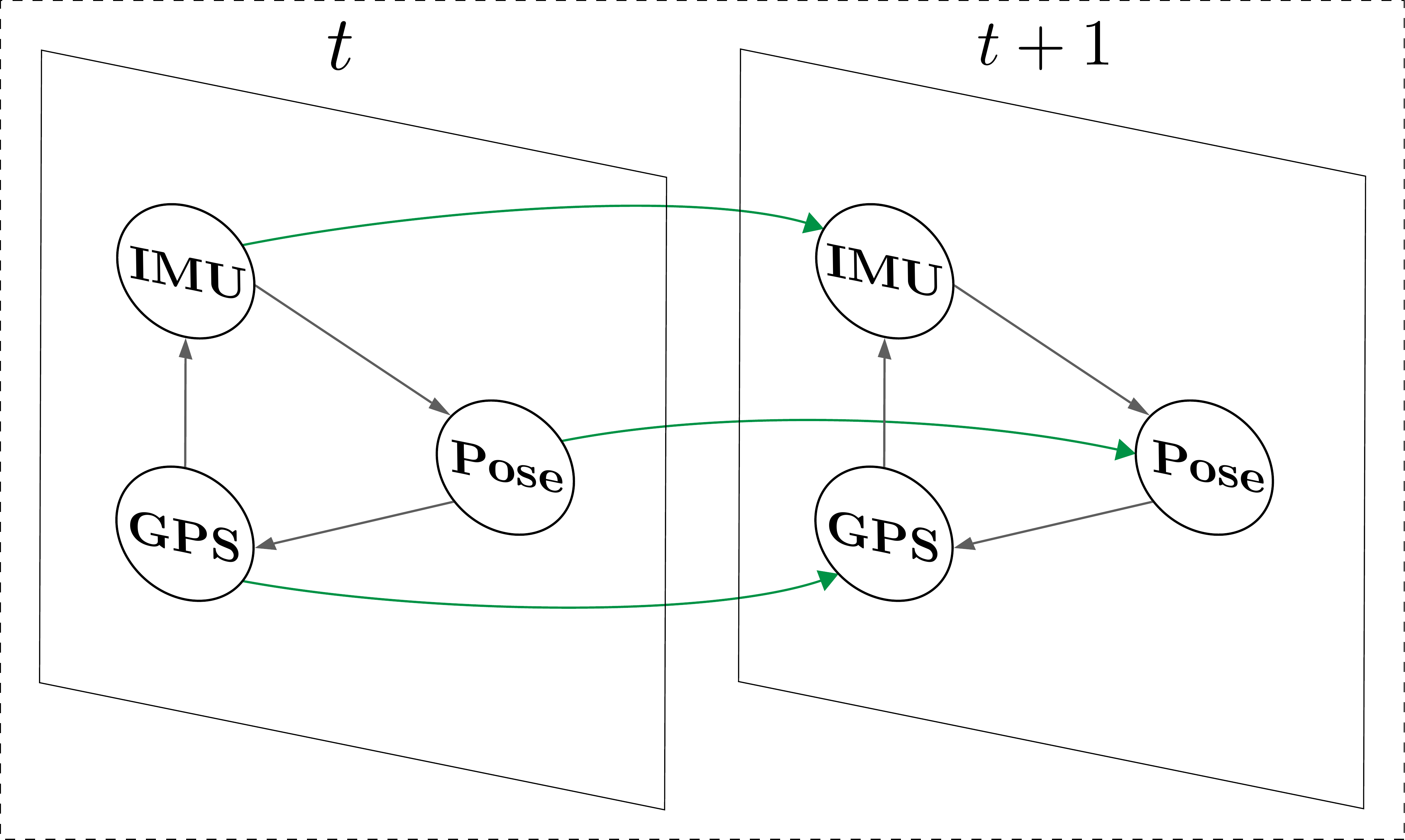}};
        \draw[text=black] (-1.85, -1.1) node {\footnotesize(b)};
        \end{tikzpicture}
  		\end{minipage}
  	\end{tabular}
  \end{minipage}
	\caption{ {\bf Temporal diagnostic graphs.} (a) Object detection system, and  (b) Vehicle localization system.} 
  \label{fig:apollo_diagnostic_graph}
\end{center}
\vspace{-9mm} 
\end{figure}

\myParagraph{Implementation details}
\rev{The Lincoln MKZ car is equipped with a GPS, IMU, RADAR, LIDAR (32 channels), a front-facing monocular camera used for obstacle detection and a telephoto camera (pointing slightly upward) used only for traffic light detection.}
The \apollo perception system has been configured to use RADAR, LIDAR, and the \rev{front-facing} camera to detect objects.
\apollo's object detection system outputs a list of objects for each sensor and then it fuses each individual detection into a single estimate using a configurable sensor fusion algorithm~\cite{apollo-auto-repo}.
\rev{In the case of RADAR and camera, each object is parametrized with its 3D bounding box.
In the case of the LIDAR each object is represented by its 3D convex hull.}
We denote with $\obs{C}$, $\obs{L}$, and $\obs{R}$ the obstacles detected by the camera, LIDAR, and RADAR, respectively, 
and with $\obs{F}$ the obstacle as determined by the sensor fusion \module~\rev{(represented by convex hulls)}.
\rev{From \cref{thm:pcm} we know that a diagnostic graph with \num{4} nodes can be up to $1$-diagnosable;  
therefore, we implemented several consistency \tests, which are visualized in~\cref{fig:apollo_diagnostic_graph}a, to achieve a $1$-diagnosable graph.}
\rev{A formal description of} each test is given in \suppMaterialLong, while here we provide a high-level overview.
For a consistency check $A\to B$ (node $A$ testing node $B$), we return $1$ (fail) if there exists an object in $A$ within the field-of-view of $B$ that is not matched by any object in $B$ or vice versa, or $0$ (pass) otherwise. 
We also add temporal \tests: for each time $\myt$, we test outputs at consecutive timestamps \ie $A_{\myt-1}\to A_{\myt}$ (consecutive object detections). When possible, we also cross-check executions of different \modules at nearby timestamps, \ie $A_{\myt-1}\to B_{\myt}$ (consecutive detections across different sensors).  
The result is the $2$-diagnosable graph shown in~\cref{fig:apollo_diagnostic_graph}a. 
Note that if we neglect the time dimension, the graph becomes $1$-diagnosable, remarking the importance of our extension in Section~\ref{sec:tdg}.
\rev{Due to the asynchronous nature of the perception system, a new diagnostic graph is created as soon as all the nodes in the graph are available, refreshing old values when new ones are available, and resulting in a new diagnostic graph every \num{0.3}-\SI{0.5}{\milli\second}}.
In our implementation, every consistency \test also returns a severity level. 
For object detection, we set this value to \emph{low} if there is an inconsistency for an object on the road (but not in the current lane), and \emph{high} if there is an inconsistency on the same lane occupied by the AV (\cref{fig:front_page}c).


\myParagraph{Scenario}
In the \lgsvl simulator we developed a scenario where an overturned truck occupies a lane in the Borregas Avenue map (\cref{fig:front_page}b).
This scenario is designed based on the recent Tesla accident in Taiwan~\cite{teslaAccidentTruck}, shown in~\cref{fig:front_page}a.
We tested the system in four different variations of the scenario:
  (i) \emph{Simple} scenario takes place at noon with clear weather and no traffic, 
  (ii) \emph{Rain} takes place at noon but with moderate rain, fog and no traffic, 
  (iii) \emph{Night} takes places at 9pm with clear weather and no traffic,
  (iv) \emph{Traffic} takes place at noon with clear weather and incoming traffic in neighboring lanes.
We average results over \num{10} random instances of each scenario, \num{5} with the truck occupying a lane and \num{5} without.
 Traffic and truck position are randomized, but the {truck (when present) always blocks the 
lane the AV is driving on.}

\myParagraph{Results}
In our tests, \apollo's perception system failed to detect the truck in 70\% of the cases, leading to \num{14} collisions.
\cref{fig:front_page}c shows the top view of a sample trajectory and also visualizes \framework's fault detections.
The red sections represent points where a high severity fault was detected, 
corresponding to the case were the truck was blocking the road but the camera failed to detect it, 
inducing failures in the sensor fusion module and resulting in a collision. 
By detecting the failure, \framework~\rev{would be able} to trigger an emergency stop and prevent the accident. 


\setlength{\tabcolsep}{1.1mm}
\renewcommand{\arraystretch}{1}
\newcommand{\specialcellbold}[2][c]{%
  \bfseries
  \begin{tabular}[#1]{@{}c@{}}#2\end{tabular}%
}
\begin{table}[hb]
  \vspace{-4mm}
  \centering
  \begin{tabular}{lcccc}
  \toprule
  \textbf{Scenario} &
  \specialcellbold{False non\\critical (\%)} &
  \specialcellbold{False\\critical (\%)} &
  \specialcellbold{Correct\\critical (\%)} &
  \specialcellbold{Avg. time to impact \\from failure (s)} \\
  \midrule
  Simple  & 43 & 0 & 100 & 4.17 \\
  Night   & 4.44  & 0.58 & 99.42 & 3.65 \\
  Rain    & 49.55 & 8.47 & 91.53 & 4.04 \\
  Traffic & 67.67 & 5.04 & 94.96 & 4.11 \\
  \bottomrule
\end{tabular}
\vspace{-1mm}
\caption{Object detection results aggregated by scenario.}
\vspace{-5mm}
\label{tab:scenario_results}
\end{table}


\cref{tab:scenario_results}  shows statistics describing \framework's effectiveness in detecting failures.
The ``correct critical'' column indicates the percentage of case 
where \framework detected high severity faults, and using it would have prevented an imminent collision (within 20 seconds). 
We observe that \framework would have prevented nearly all accidents in our tests.
The ``false non-critical'' column reports the percentage of cases where 
 \framework triggered low severity faults that did not resolve in a collisions; 
these are typically caused by the \apollo camera-based object detection, which 
mis-detected obstacles on the sidewalks or other lanes. 
These are frequent in modern perception systems, and are classified as non critical by \framework. 
The ``Traffic'' scenario exhibits the highest number of false non-critical faults: they are the result of the camera failures in the proximity of an incoming vehicle.
The ''false critical'' column describes the percentage of tests where \framework would have stopped the 
car even if no accident was imminent; this percentage is low and can be further improved upon (Section~\ref{sec:conclusions}).
The ``Rain'' scenario has the largest number of false critical alarms, because in few occasions \apollo's perception system detected regions of the wet road 
floor as obstacles, hence triggering a fault detection and incorrectly stopping the vehicle.
The experiment shows that even if \framework may be conservative, it reliably reports high severity faults.
\rev{In \cref{sec:conclusions} we discuss possible extensions to this work that can better cope with false alarms.}

\subsubsection{Vehicle localization}
In this section, we use \framework to detect failures in \apollo's AV localization system. 

\myParagraph{Implementation details}
\rev{The car moves at a maximum speed of \SI{45}{\mph} and a default cruise speed of \SI{25}{\mph}.} The \apollo vehicle localization system is configured to use the GNSS RTK module.
We denote with GPS and IMU the output of the two sensors while we denote as ``Pose Estimate'' the estimate produced by \apollo's sensor fusion algorithm.
We implemented \num{6} consistency \tests, \num{3} regular consistency \tests and \num{3} temporal consistency \tests. 
The pose estimate and the GPS measurement are compared directly and the \test returns $1$ (fail) if the difference in the positions is above \SI{1}{\meter} threshold.
The comparison with the IMU verifies that the change in the pose estimate is compatible with the measured acceleration and angular velocities {(Input/Output Consistency)} and within the physical limits of the car {(Input Admissibility)}.
As temporal \tests, 
we compare the output of each \module at consecutive time instants, \ie $A_{\myt-1}\to A_{\myt}$.
The resulting  temporal diagnostic graph is $1$-diagnosable and visualized in~\cref{fig:apollo_diagnostic_graph}b.
A new temporal diagnostic graph is built every \SI{0.1}{\milli\second} and we do not distinguish the faults by severity.

\myParagraph{Scenario}
To test the localization system we used the same scenarios of the previous section, 
but randomly spoiled GPS measurements with incorrect data.
The duration of the fault was randomized to last between \num{0.1} and \num{0.4} seconds.

\begin{figure}[ht!]
  \centering
  \vspace{-2mm}
  \includegraphics[width=\linewidth]{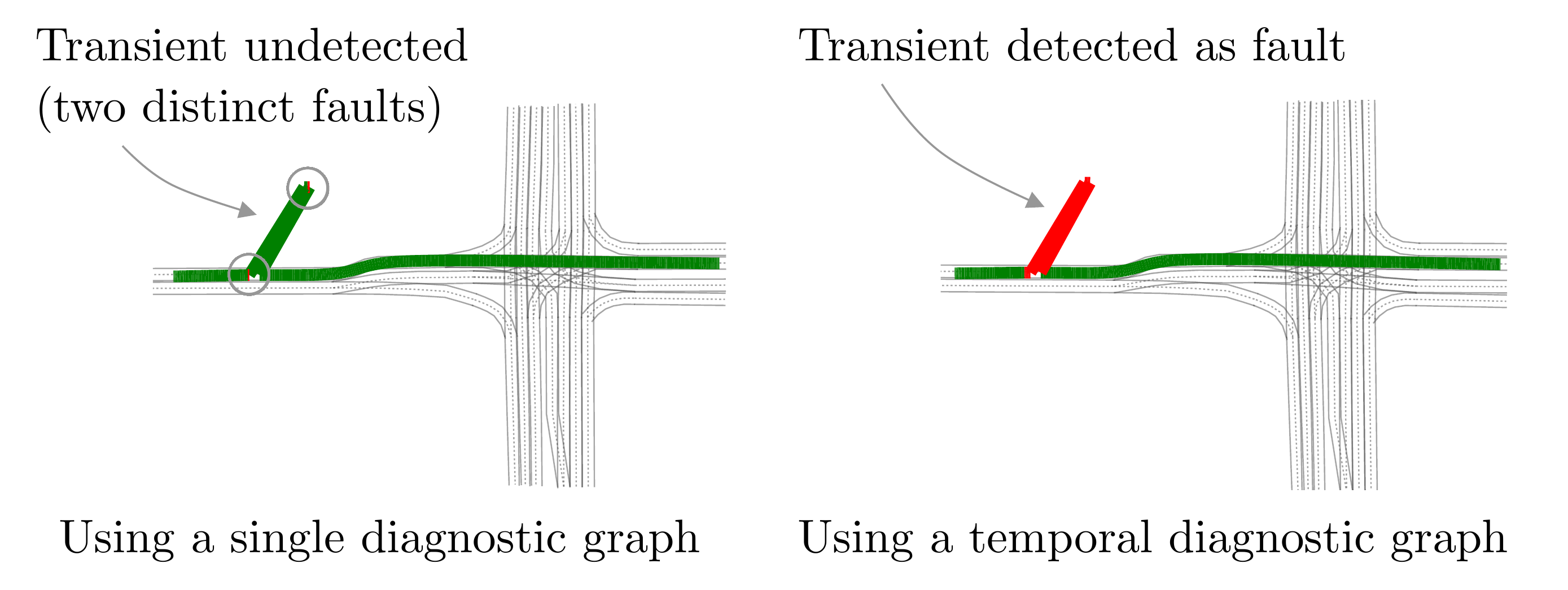}
  \vspace{-8mm}
  \caption{{\bf Vehicle localization monitoring.} Comparison between diagnostic graph and temporal diagnostic graph for vehicle localization (red: \framework detected a fault, green: \framework did not detect any faults).   \label{fig:gps_fault_comparison}
  \vspace{-4mm}}
\end{figure}

\myParagraph{Results}
In all the tests, \apollo was able to stop the car without accidents in response to the incorrect GPS readings.
We tested \framework with a single diagnostic graph and with a temporal diagnostic graph made by \num{2} consecutive 
diagnostic graphs (\cref{fig:apollo_diagnostic_graph}b).
\framework was also able to detect the failures in all tests, and it correctly identified the GPS and the Pose Estimate 
as faulty \num{46.43}\% of the times in the case of the temporal diagnostic graphs (\cref{fig:apollo_diagnostic_graph}b).
In the case of a single diagnostic graph, \framework failed to identify the faulty modules \num{100}\% of the times, 
remarking the importance of using temporal information for monitoring.
Incorrect fault identification is due to the fact that when both the GPS and the sensor fusion fail, 
we end up with two faults in a $1$-diagnosable temporal graph, which cannot be uniquely identified.
\rev{By equipping the sensor fusion with an outlier detector, 
it would be possible to produce a correct pose estimation using only the IMU;
 in such a case, the system would have experienced one fault in a 
$1$-diagnosable graph, making fault identification possible.
}
\cref{fig:gps_fault_comparison} overlays the output of \framework on the vehicle trajectory, comparing 
traditional and temporal diagnostic graphs.
\rev{This example strongly indicates the need of using additional sources of information, like visual-inertial odometry, 
to increase redundacy in the system and hence increase diagnosability.}



\section{Conclusion}\label{sec:conclusions}

We presented \framework, a novel framework for fault diagnosis in perception systems.  
\framework reasons over the consistency of the perception results across \modules and over time and is shown to 
be able to  detect failures (and prevent accidents) in state-of-the-art perception systems.
\framework builds on the distant literature of multiprocessor diagnosability, 
which we extend and tailor to modern perception systems. 
In particular, we borrow the notion of \emph{diagnosability}, which can be understood as a measure of robustness and 
quantifies the maximum number of faults that can be correctly identified in a system.  
Moreover, we extend existing work and introduce \emph{temporal diagnostic graphs}, which describe perception \modules 
interacting at different rates and over time, while still enabling the use of existing fault identification algorithms. 

We believe that \emph{temporal diagnostic graphs} can complement the literature and enhance the current 
practice, contributing to the goal of achieving safe and trustworthy autonomous vehicles. 
For instance, a system designer can use the tools proposed in this paper to assess the diagnosability before deploying 
the vehicle on public roads, or design the system in such a way that its diagnosability is maximized.
At runtime, \framework allows the vehicle to have a greater awareness of its operational envelope and 
enables real-time perception monitoring; 
\framework's consistency tests formalize and generalize watchdogs used in standard practice.
For regulators, a certain level of diagnosability can be used as a requirement for certification, and, 
in the unfortunate case of an accident, the proposed approach may increase accountability by providing
formal evidence about the root cause of the failures. 

In future work, we plan to extend \framework to be a probabilistic framework that  
accounts for, and reasons over, the probability of failure. 
\rev{Such a framework would potentially allow coping with missing (due to asynchronous updates) and unreliable tests, and better model fault propagation.}

\setcounter{equation}{0}
\setcounter{section}{0}
\setcounter{theorem}{0}
\setcounter{figure}{0}

\bibliographystyle{ieeeTran}
\bibliography{../../references/refs.bib, ../../references/myRefs.bib}

\isExtended{
  \appendices 
  \section{Consistency tests}
\label{sec:consistencyTests}

We list all the consistency tests used in the experimental evaluation with \apollo.

\subsection{Notation}

\begin{definition}[Approximate set membership]
With $ x \approxin S$ we denote the approximate set membership.
We assume there exist a distance function $\dist_S(\cdot, \cdot)$ and a given threshold $\tau_S$ such that, if $x\approxin S$ we have
\begin{equation*}
    \exists y \in S \st \dist_S(x,y) \leq \tau_S
\end{equation*}
\end{definition}

When we use the approximate set membership to determine if two obstacles are approximately the same, the distance function depends on the data structure representing them while the threshold $\tau_S$ depends on their size.

\begin{definition}[Objects in Field of View]
Given a sensor $A$ we denote with $\fov(A)$ the set of obstacles that fall in the field of view of the sensor.
\end{definition}

\begin{definition}[Map and Current Lane]
  We denote with $\mathrm{Map}$ the set of points that belong to the map. 
  Similarly, we denote with $\mathrm{CurrentLane}$ the set of points that belong to the lane currently occupied by the AV. 
\end{definition}

When determining if an obstacle belongs to the Field of View of a sensor and/or a lane, we use the approximate set membership operator.
Moreover, we denote:
\begin{itemize}
  \item $\obs{R}$ the set of objects detected by the RADAR
  \item $\obs{L}$ the set of objects detected by the LIDAR
  \item $\obs{C}$ the set of objects detected by the camera
  \item $\obs{F}$ the set of objects returned by the sensor fusion alg.
  \item $\tau_p$ the maximum position error allowed (\SI{30}{\cm})
  \item $\tau_v$ the maximum velocity error allowed (\SI{0.5}{\meter/\second})
  \item $\hat{v}$ the maximum vehicle speed (\SI{90}{\kilo\meter/\hour})
  \item $\hat{a}$ the maximum vehicle acceleration (\SI{10}{\meter/\second^2})
  \item $\hat{j}$ the maximum vehicle jerk (\SI{15}{\meter/\second^3})
\end{itemize}
\subsection{Consistency tests for object detection}

\newcommand{\percct}[2]{
  \ct{$\obs{{#1}}\to\obs{{#2}}$}{
  \begin{math}
    \begin{aligned}
      &U_{#1}^{#2} = \{o\in\obs{{#1}} \>|\> o\in\fov({#2}) \land o \not\approxin\obs{{#2}}\} \\
      &U_{#2}^{#1} = \{o\in\obs{{#2}} \>|\> o\in\fov({#1}) \land o \not\approxin\obs{{#1}}\} \\
      &U = U_{#1}^{#2} \cup U_{#2}^{#1}  \\
      &\textbf{return } |U| > 0,\> \mathrm{Severity}(U)
    \end{aligned}
  \end{math}
}
}

\percct{R}{F}
\vspace{-0.5em}
\percct{F}{L}
\vspace{-0.5em}
\percct{L}{R}
\vspace{-0.5em}
\percct{R}{C}
\FloatBarrier

The set $U_A^B$ represents the object detected by $A$ that are in $B$ field-of-view but we were unable to match (thus unmatched).
The set $U = U_A^B \cup U_B^A$ represents the the set object that one of the two sensors detected but the other did not.
If the set $U$ is non-empty, the consistency test reports a fault.
The severity is computed as follow
\begin{equation*}
  \mathrm{Severity}(U) = \begin{cases} 
    \text{HIGH} &\mbox{if } \exists o \in U \st o \in \mathrm{CurrentLane} \\ 
    \text{LOW}  &\mbox{if } \exists o \in U \st o \in \mathrm{Map}\setminus\mathrm{CurrentLane} \\
    \text{NONE} &\text{otherwise}
  \end{cases}
\end{equation*}

As temporal consistency tests we use the same consistency tests we described above but with the proper change of inputs.

\subsection{Consistency tests for vehicle localization}

Given $p_{t_1},v_{t_1}$ and $p_{t_2},v_{t_2}$ the position and velocity estimated by the pose estimation \module at the time $t_1$ and $t_2$, and the set of acceleration measurements  $a_i\ldots a_j$ collected by the IMU in the interval $[t_1,t_2]$:
\ct{IMU $\to$ POSE}{ 
  \begin{math}
    \begin{aligned}
      r =\> &|p_{t_2}-p_{t_1}|\leq \hat{v}(t_{t_2}-{t_1}) \land \ldots \\
      &|(v_{t_2}-v_{t_1}) - \int_{t_1}^{t_2} a(t) dt | \leq \tau_v \land \ldots \\
      &a_i,\ldots a_j \leq \hat{a} \land v_{t_1} \leq \hat{v} \land v_{t_2}\leq \hat{v}\\
      &\hspace{-1.8em}\textbf{return } \lnot r
    \end{aligned}
  \end{math}  
  \vspace{-3mm}
}
\FloatBarrier

Given $p_e,v_e$ the position and velocity estimated by the pose estimation \module , and the the position and velocity $p_g,v_g$ measured by the GPS:
\ct{POSE $\to$ GPS}{ 
  \begin{math}
    \begin{aligned}
      r =\> &|p_e-p_g|\leq \tau_p \land |v_p-v_g|\leq\tau_v \land \ldots \\
      &v_e \leq \hat{v} \land v_g\leq \hat{v}\\
      &\hspace{-1.8em}\textbf{return } \lnot r
    \end{aligned}
  \end{math}  
}
\FloatBarrier

Given $p_{t_1},v_{t_1}$ and $p_{t_2},v_{t_2}$ the position and velocity measured by the GPS, and the set of acceleration measurements  $a_i\ldots a_j$ collected by the IMU in the interval $[t_1,t_2]$:
\ct{GPS $\to$ IMU}{ 
  \begin{math}
    \begin{aligned}
      r =\> &|p_{t_2}-p_{t_1}|\leq \hat{v}(t_{t_2}-{t_1}) \land \ldots \\
      &|(v_{t_2}-v_{t_1}) - \int_{t_1}^{t_2} a(t) dt | \leq \tau_v \land \ldots \\
      &v_{t_1} \leq \hat{v} \land v_{t_2}\leq \hat{v}\\
      &\hspace{-1.8em}\textbf{return } \lnot r
    \end{aligned}
  \end{math}  
}
\FloatBarrier

Given $p_{t_1},v_{t_1}$ and $p_{t_2},v_{t_2}$ the position and velocity measured by the GPS at time $t_1$ and $t_2$:
\ct{GPS $\to$ GPS}{ 
  \begin{math}
    \begin{aligned}
      r =\> &|p_{t_2}-p_{t_1}|\leq \tau_p \land |v_{t_2}-v_{t_1}|\leq\tau_v \land \ldots \\
      &v_{t_1} \leq \hat{v} \land v_{t_2}\leq \hat{v}\\
      &\hspace{-1.8em}\textbf{return } \lnot r
    \end{aligned}
  \end{math}
}
\FloatBarrier

Similarly, given $p_{t_1},v_{t_1}$ and $p_{t_2},v_{t_2}$ the position and velocity estimated by the pose estimation filter at time $t_1$ and $t_2$:
\ct{POSE $\to$ POSE}{ 
  \begin{math}
    \begin{aligned}
      r =\> &|p_{t_2}-p_{t_1}|\leq \tau_p \land |v_{t_2}-v_{t_1}|\leq\tau_v \land \ldots \\
      &v_{t_1} \leq \hat{v} \land v_{t_2}\leq \hat{v}\\
      &\hspace{-1.8em}\textbf{return } \lnot r
    \end{aligned}
  \end{math}  
}
\FloatBarrier

Finally, for the IMU we limit the jerk required to change the measured accelerations.
Given two acceleration measurements $a_{t_1}$ and $a_{t_2}$:
\ct{IMU $\to$ IMU}{ 
  \begin{math}
    \begin{aligned}
      r =\> &\frac{a_{t_2}-a_{t_1}}{t_2-t_1}\leq\tau_j \land \ldots\\
      &a_{t_1}\leq \tau_a \land a_{t_2}\leq \tau_a \\
      &\hspace{-1.8em}\textbf{return } \lnot r
    \end{aligned}
  \end{math}  
}
\FloatBarrier

\section{Proof}
\label{sec:proof}

\textbf{Proof of \cref{cor:more_edges}.}
Assume conditions (i), (ii), and (iii) in \cref{thm:pcm} hold for the pair $\pair{D,\maxFaults}$.
Since $D$ and $D'$ have the same number of nodes, condition (i) holds for $\pair{D',\maxFaults}$.
Since $\delta_{\mathrm{in}}(D)\leq\delta_{\mathrm{in}}(D')$ and for every $X\ss U$ we have $\Gamma_D(X)\leq\Gamma_{D'}(X)$, conditions (ii) and (iii) also hold for $\pair{D',t}$, proving the claim.

\textbf{Proof of \cref{prop:subgraph_diag}.}
Assume each $D_i$ is $\maxFaults$-diagnosable, and suppose that $\sigma$ is a syndrome for a $\maxFaults$-consistent fault set on 
$D$ (so it has at most $\maxFaults$-many faulty nodes).
Then its restriction to each $D_i$ has fewer than $\maxFaults$-many faulty nodes, so the faulty nodes in $D_i$ can be accurately diagnosed.
Since every node is in some $D_i$, every node can be accurately diagnosed, so $D$ is $\maxFaults$-diagnosable.


\vfill\null
\columnbreak
\section{Extra Results}
\label{sec:extraResults}


\begin{figure}[ht!]
  \vspace{-1mm}
	\begin{center}
	\begin{minipage}{\textwidth}
	\begin{tabular}{c}%
		\begin{minipage}{\mpColTwo}%
        \centering%
        \begin{tikzpicture}
        \draw (0, 0) node[inner sep=0] {\includegraphics[trim=0 50 0 100,clip,width=0.92\columnwidth]{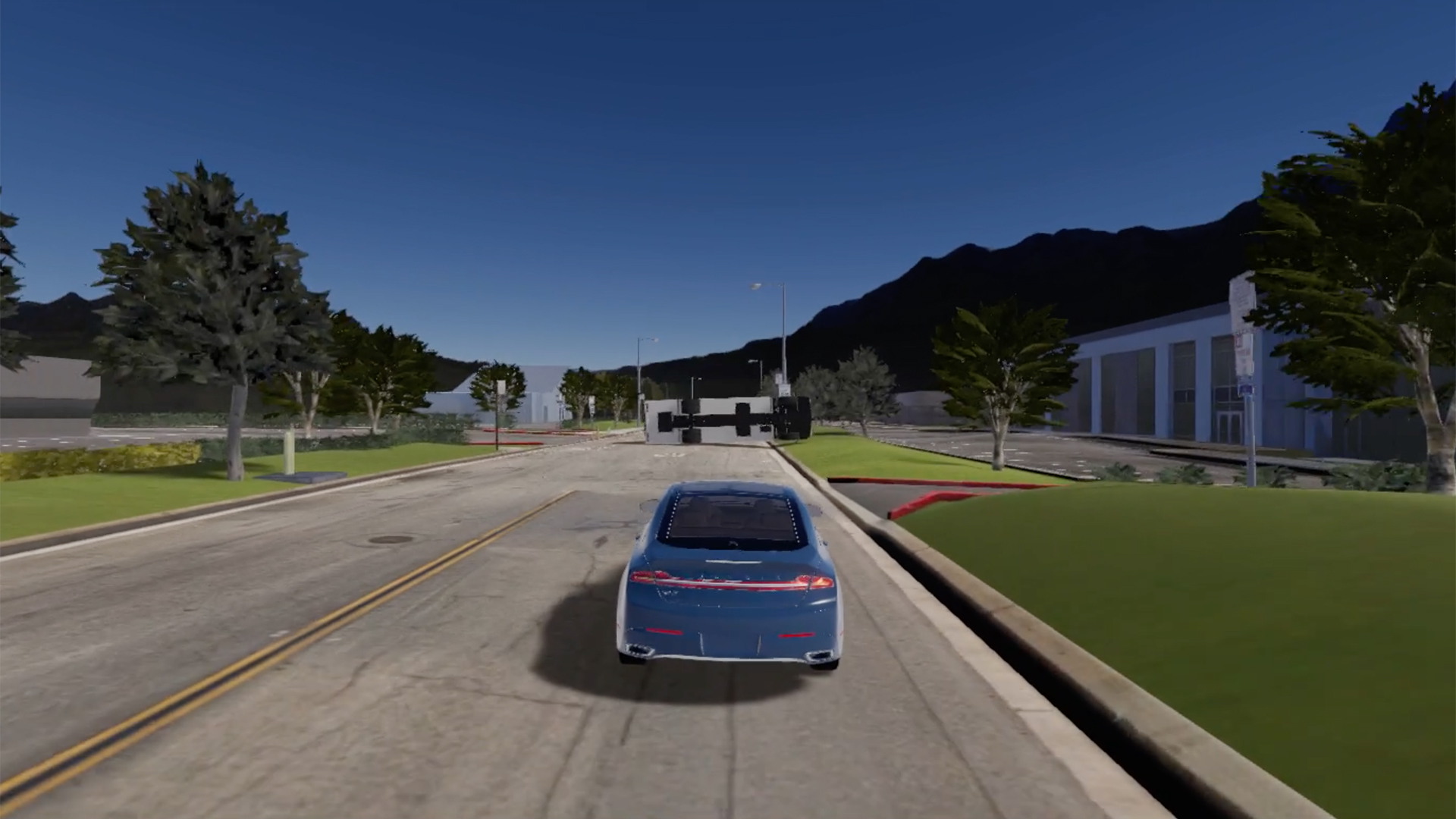}};
        \draw[text=white] (-3.6, -1.3) node {(a)};
        \end{tikzpicture}
		\end{minipage}
		\vspace{2mm}\\
		\begin{minipage}{\mpColTwo}%
        \centering%
        \begin{tikzpicture}
        \draw (0, 0) node[inner sep=0] {\includegraphics[trim=0 0 0 50,clip,width=0.92\columnwidth]{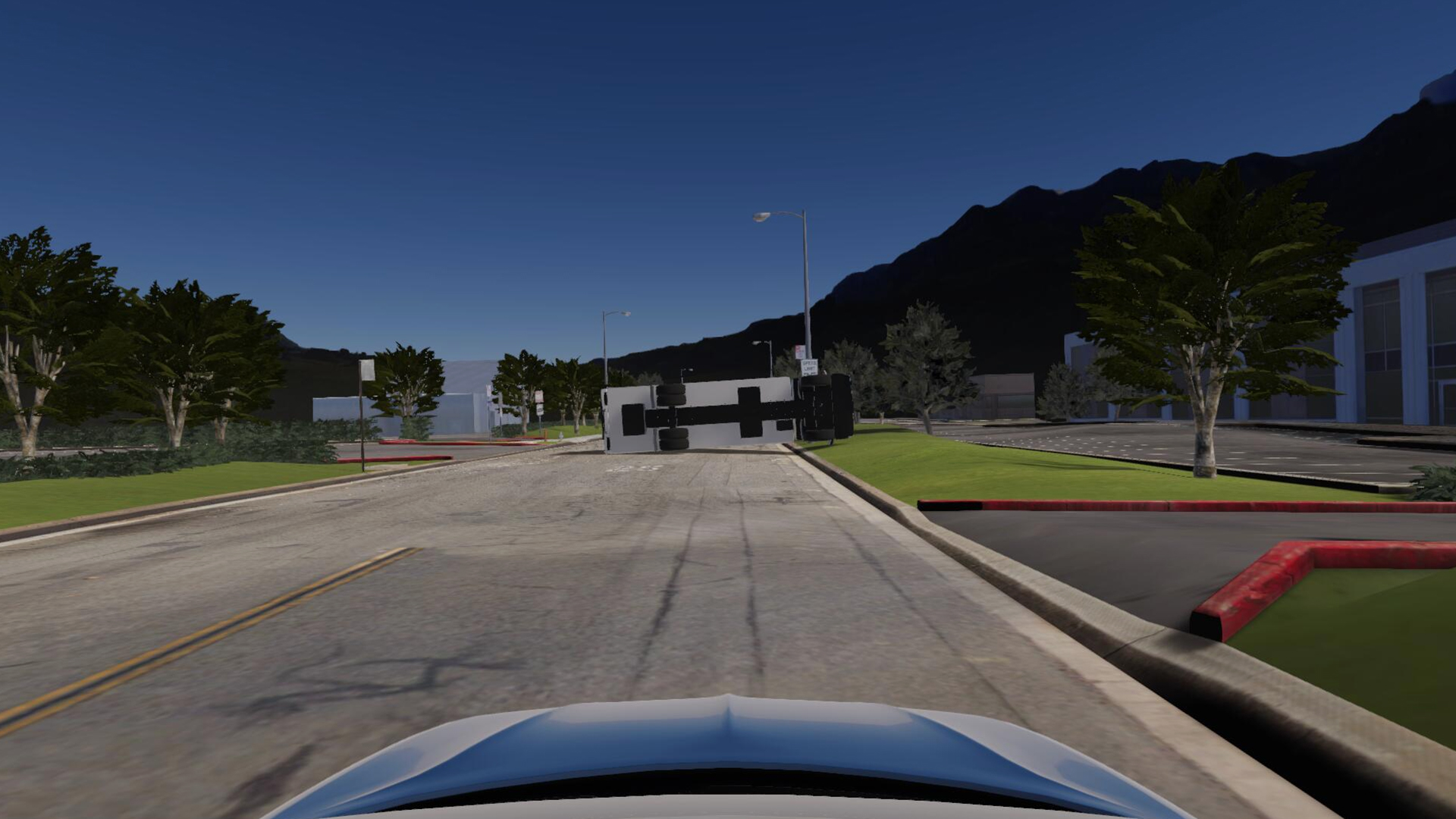}};
        \draw[text=white] (-3.6, -1.3) node {(b)};
        \end{tikzpicture}
		\end{minipage}
		\vspace{2mm}\\
		\begin{minipage}{\mpColTwo}%
        \centering%
        \begin{tikzpicture}
        \draw (0, 0) node[inner sep=0] {\includegraphics[trim=0 30 0 0,clip,width=0.92\columnwidth]{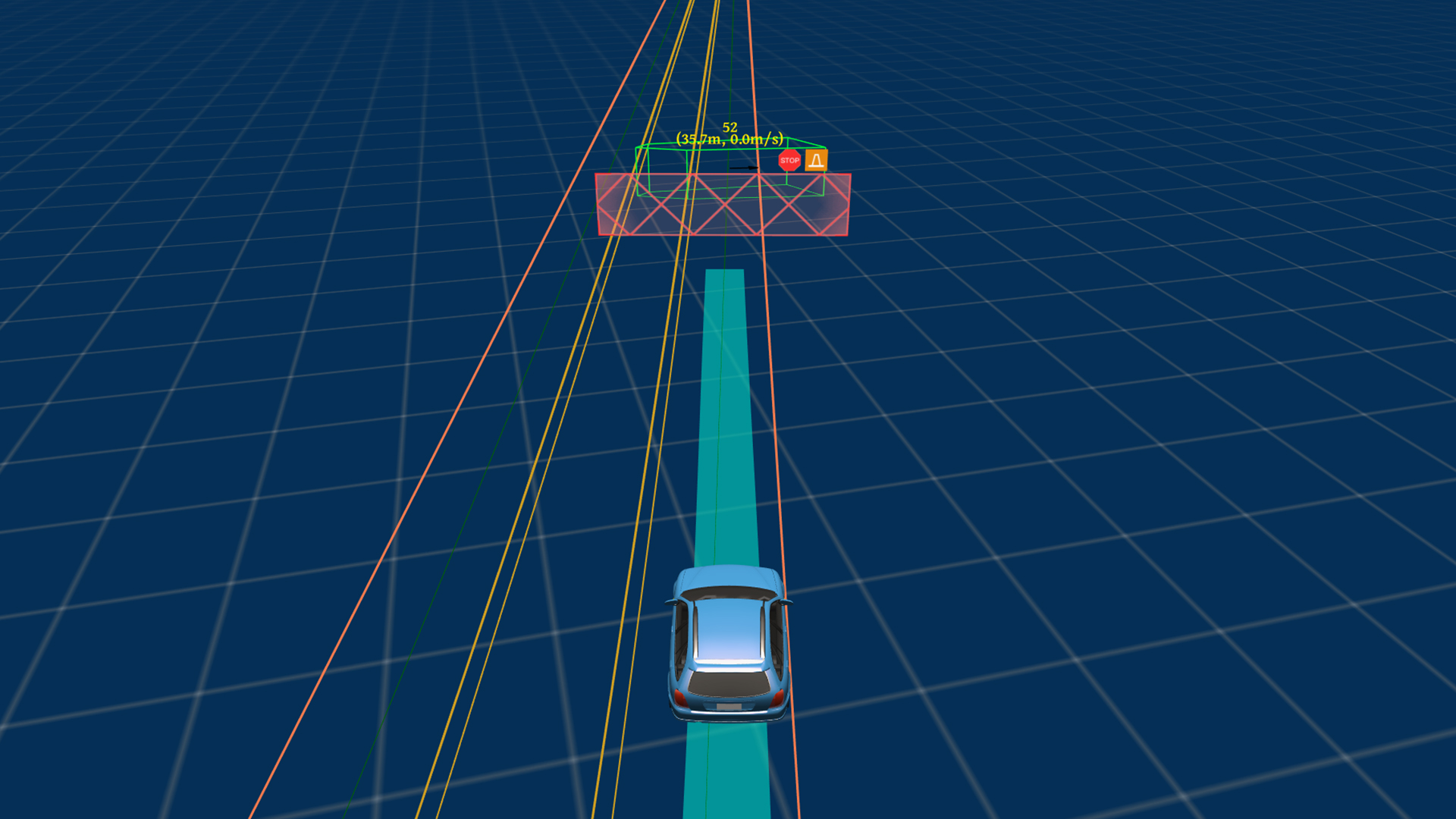}};
        \draw[text=white] (-3.6, -1.3) node {(c)};
        \end{tikzpicture}
    \end{minipage}
    \vspace{2mm}\\
		\begin{minipage}{\mpColTwo}%
        \centering%
        \begin{tikzpicture}
        \draw (0, 0) node[inner sep=0] {\includegraphics[trim=0 80 0 0,clip,width=0.92\columnwidth]{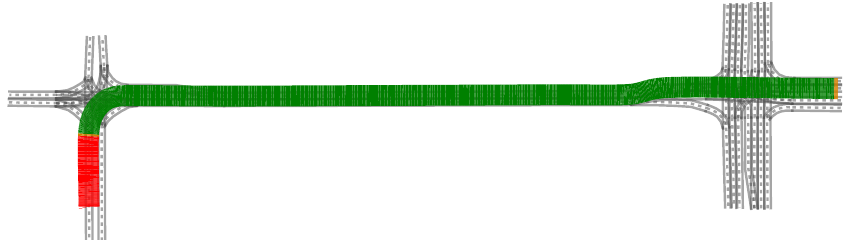}};
        \draw[text=black] (-3.6, -1) node {(d)};
        \end{tikzpicture}
		\end{minipage}
		\end{tabular}
	\end{minipage}
	\mpPostSpace
  \caption{\textbf{Simple scenario variation.} (a) Simulator, (b) Camera image, (c) World model (\apollo Dreamview), (d) output of \framework.
  \apollo sensor fusion correctly detects the overturned truck in the current lane as obstacle but the camera fails to do so, \framework is able to detect the inconsistency and report a high severity fault (red portion of the trajectory).}
	\end{center}
\end{figure}


\begin{figure}[ht!]
  \vspace{-1mm}
	\begin{center}
	\begin{minipage}{\textwidth}
	\begin{tabular}{c}%
		\begin{minipage}{\mpColTwo}%
        \centering%
        \begin{tikzpicture}
        \draw (0, 0) node[inner sep=0] {\includegraphics[width=0.95\columnwidth]{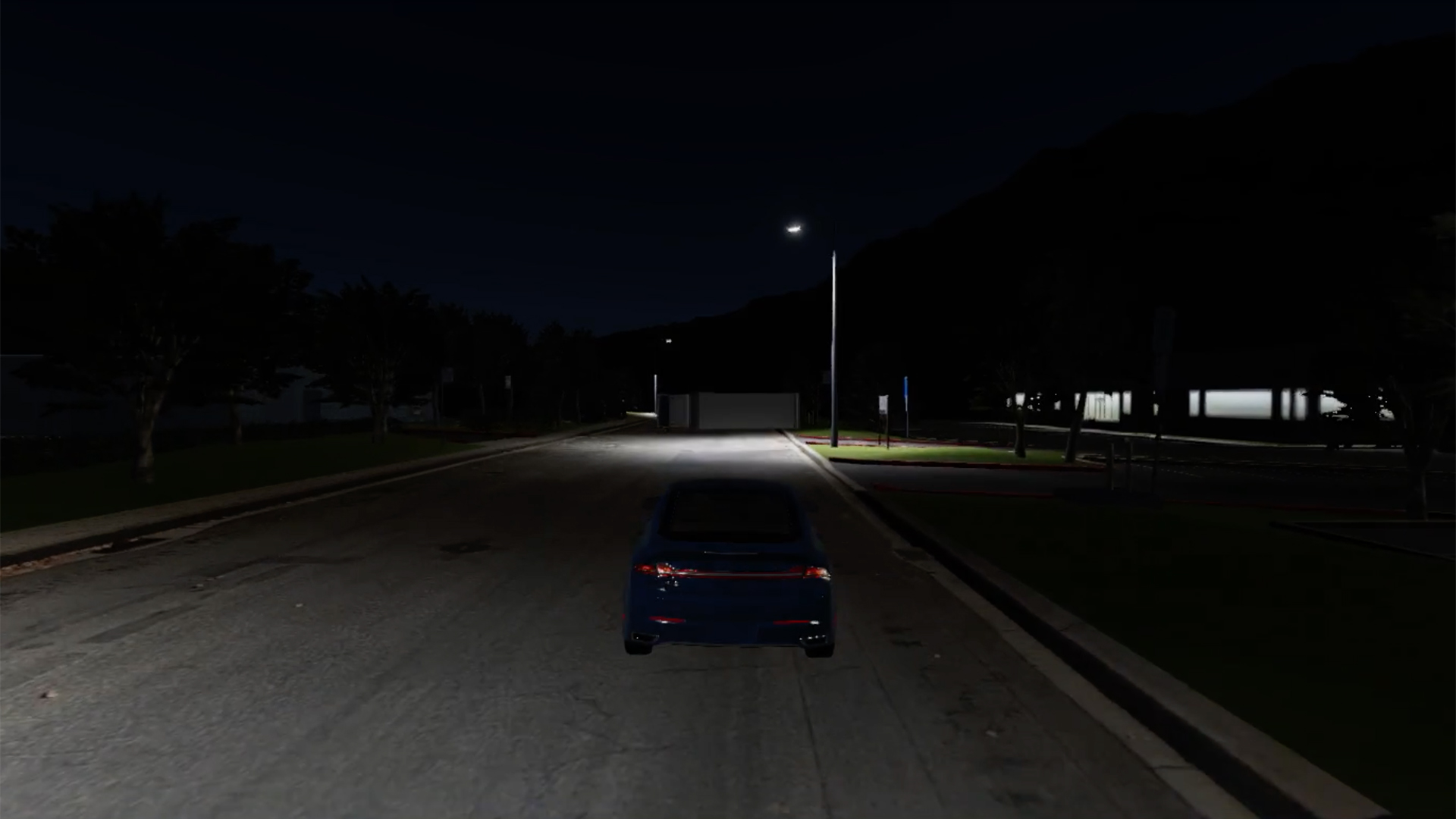}};
        \draw[text=white] (-3.8, -2) node {(a)};
        \end{tikzpicture}
		\end{minipage}
		\vspace{2mm}\\
		\begin{minipage}{\mpColTwo}%
        \centering%
        \begin{tikzpicture}
        \draw (0, 0) node[inner sep=0] {\includegraphics[width=0.95\columnwidth]{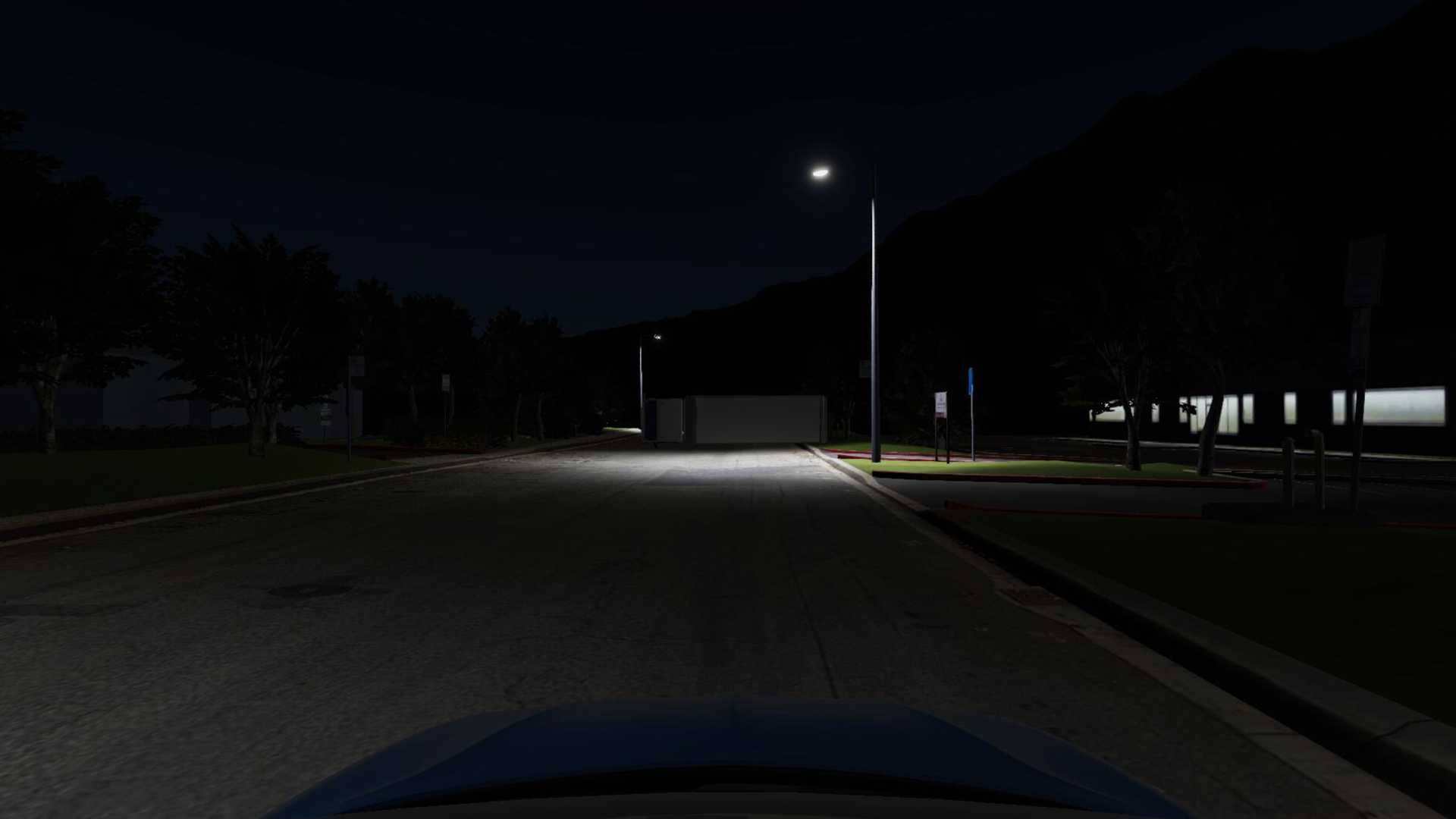} };
        \draw[text=white] (-3.8, -2) node {(b)};
        \end{tikzpicture}
		\end{minipage}
		\vspace{2mm}\\
		\begin{minipage}{\mpColTwo}%
        \centering%
        \begin{tikzpicture}
        \draw (0, 0) node[inner sep=0] {\includegraphics[width=0.95\columnwidth]{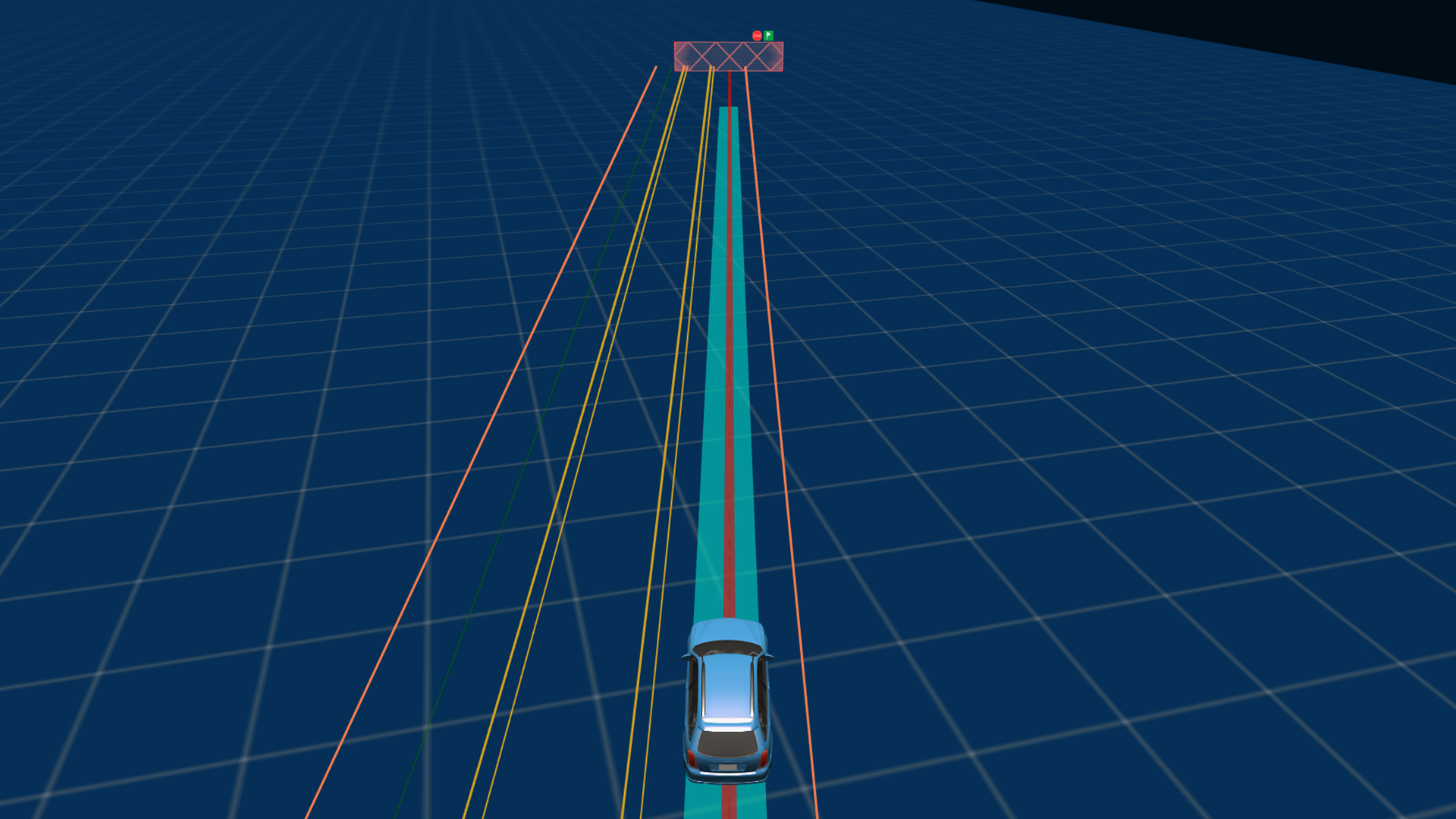}};
        \draw[text=white] (-3.8, -2) node {(c)};
        \end{tikzpicture}
    \end{minipage}
    \vspace{2mm}\\
		\begin{minipage}{\mpColTwo}%
        \centering%
        \begin{tikzpicture}
        \draw (0, 0) node[inner sep=0] {\includegraphics[width=0.95\columnwidth]{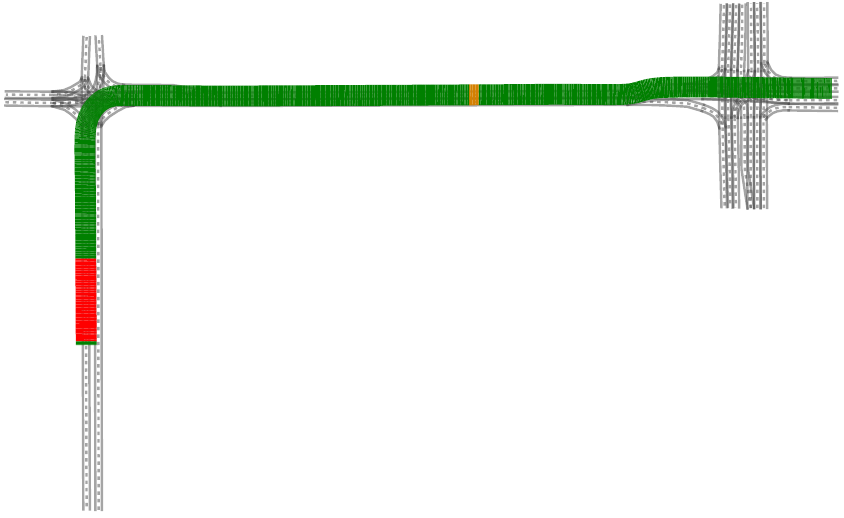}};
        \draw[text=black] (-3.8, -2) node {(d)};
        \end{tikzpicture}
		\end{minipage}
		\end{tabular}
	\end{minipage}
	\mpPostSpace
  \caption{\textbf{Night scenario variation.} (a) Simulator, (b) Camera image, (c) World model (\apollo Dreamview), (d) output of \framework.
  \apollo sensor fusion fails to detect the overturned truck in the current lane, \framework is able to detect the inconsistencies in the current lane between the radar-based obstacle detection and the other \modules and report high severity faults (red portion of the trajectory).}
	\end{center}
\end{figure}


\begin{figure}[ht!]
  \vspace{-1mm}
	\begin{center}
	\begin{minipage}{\textwidth}
	\begin{tabular}{c}%
		\begin{minipage}{\mpColTwo}%
        \centering%
        \begin{tikzpicture}
        \draw (0, 0) node[inner sep=0] {\includegraphics[width=0.95\columnwidth]{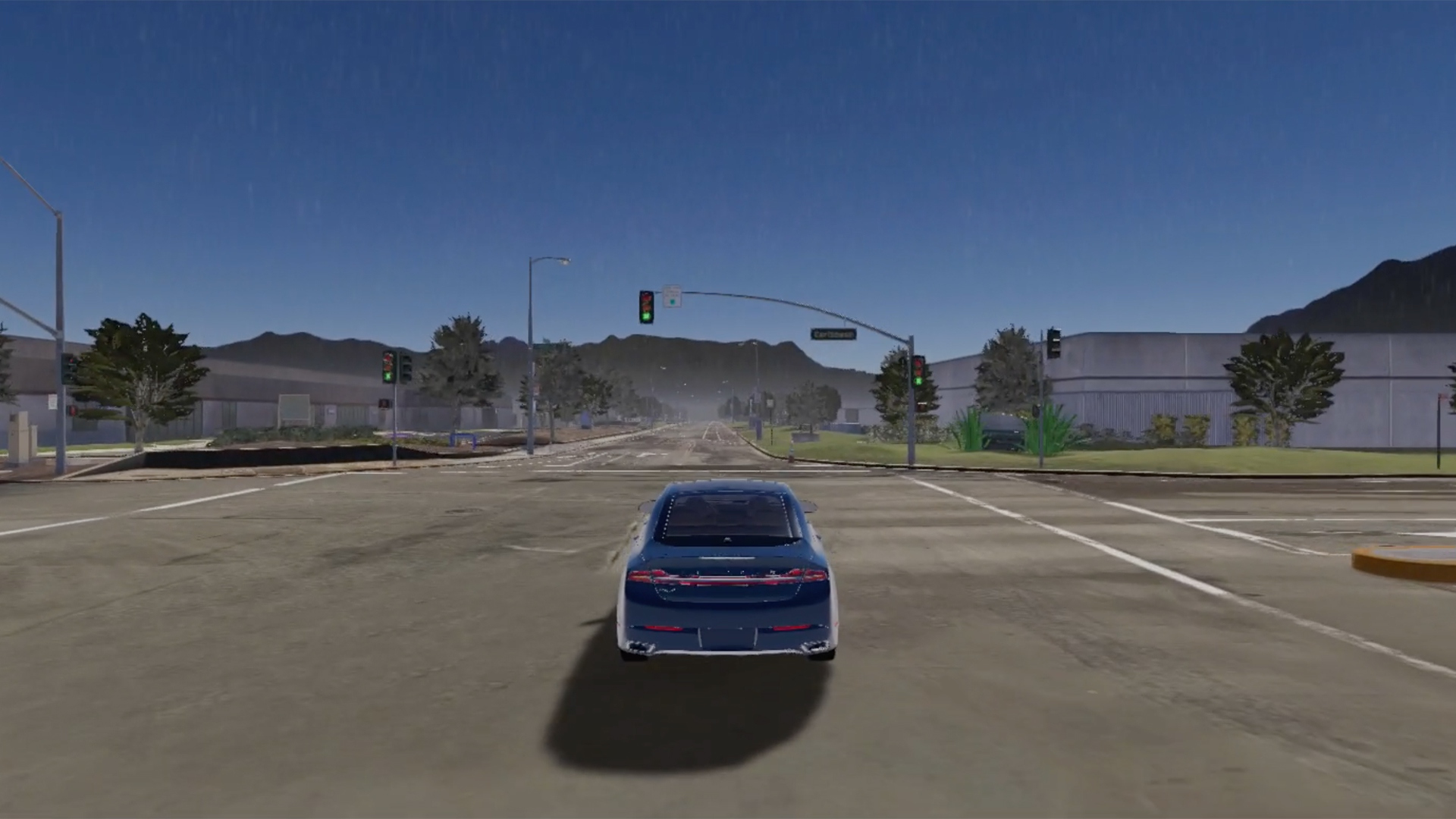} };
        \draw[text=white] (-3.8, -2) node {(a)};
        \end{tikzpicture}
		\end{minipage}
		\vspace{2mm}\\
		\begin{minipage}{\mpColTwo}%
        \centering%
        \begin{tikzpicture}
        \draw (0, 0) node[inner sep=0] {\includegraphics[width=0.95\columnwidth]{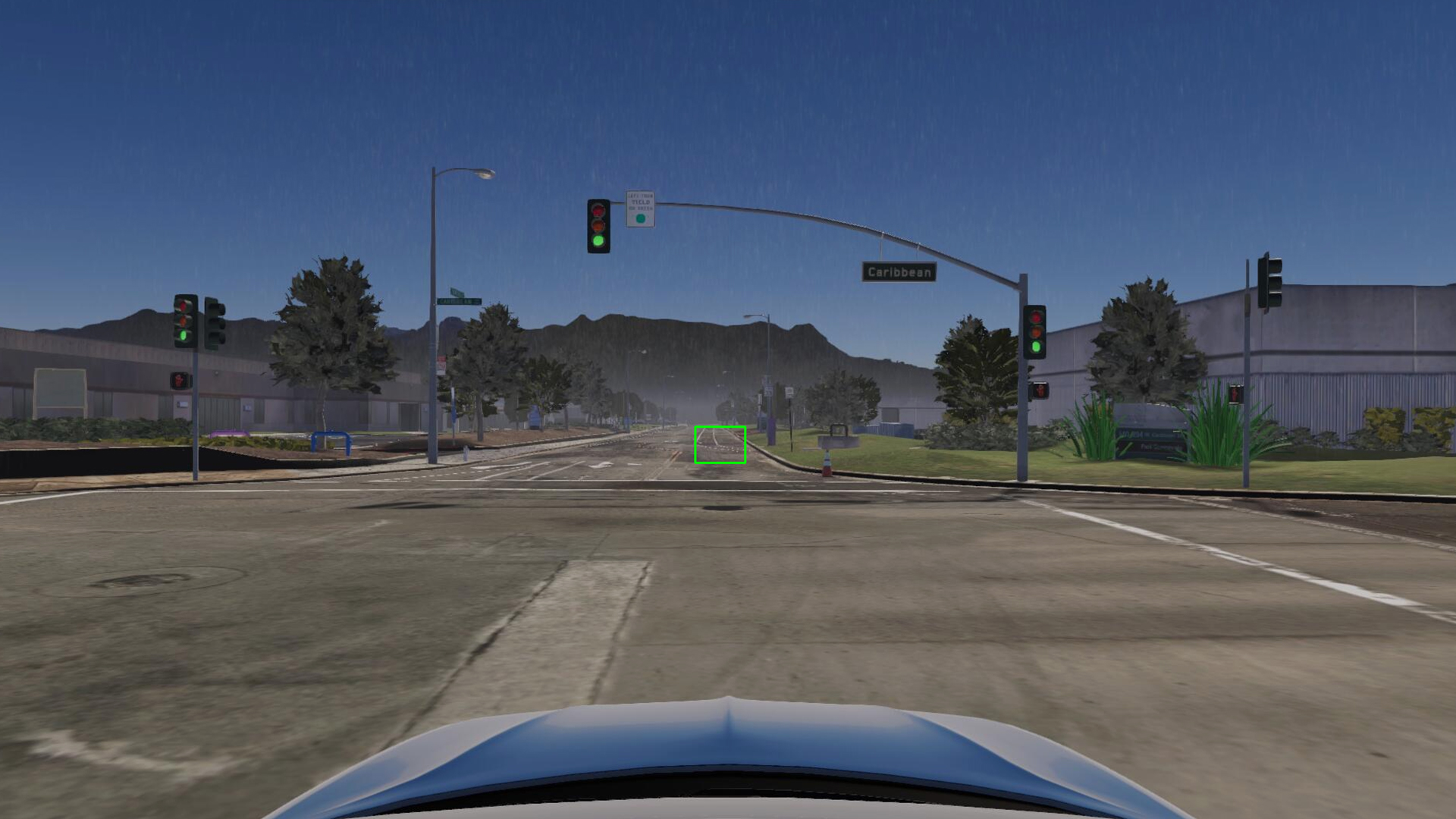}};
        \draw[text=white] (-3.8, -2) node {(b)};
        \end{tikzpicture}
		\end{minipage}
		\vspace{2mm}\\
		\begin{minipage}{\mpColTwo}%
        \centering%
        \begin{tikzpicture}
        \draw (0, 0) node[inner sep=0] {\includegraphics[width=0.95\columnwidth]{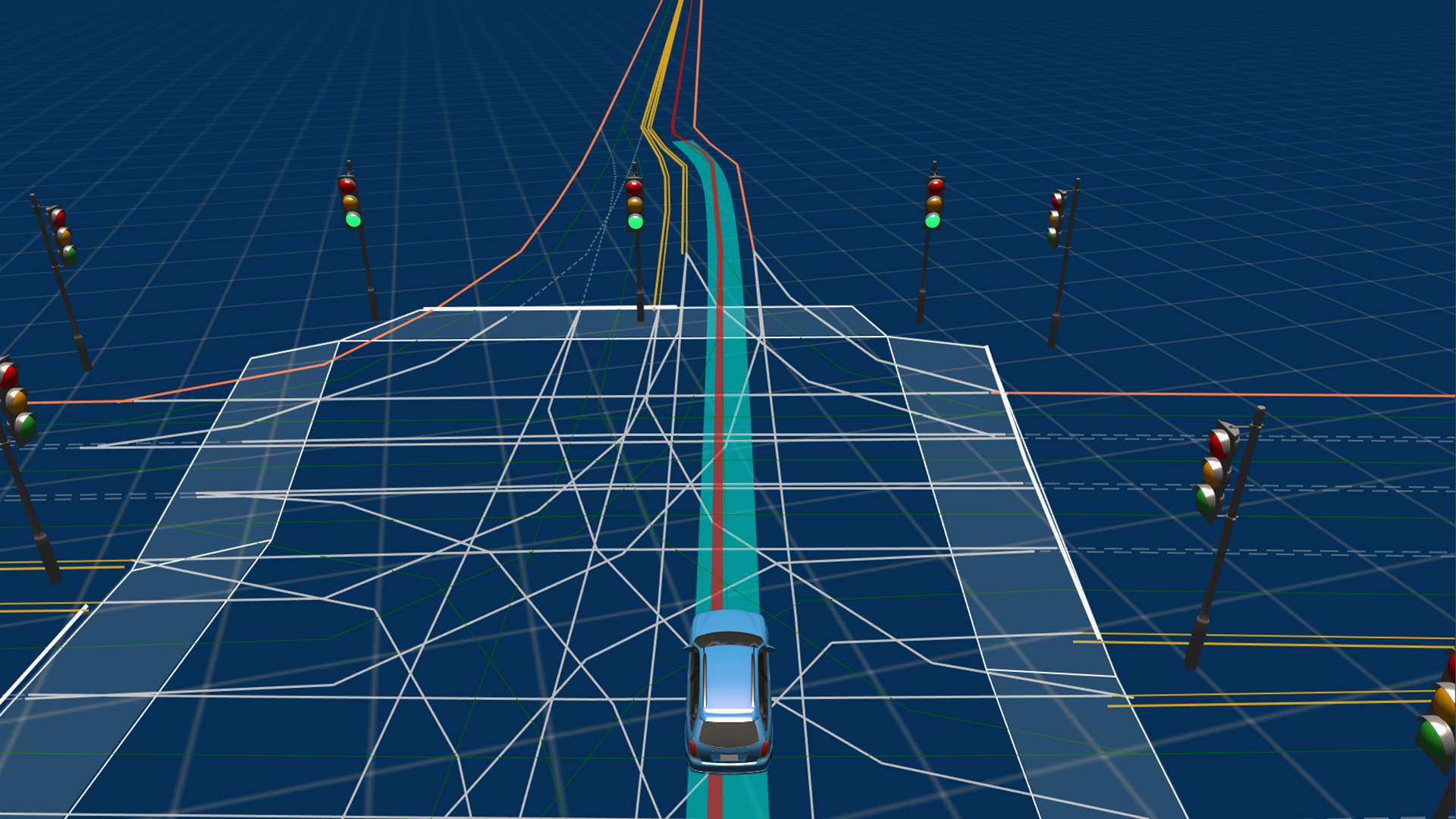} };
        \draw[text=white] (-3.8, -2) node {(c)};
        \end{tikzpicture}
    \end{minipage}
    \vspace{2mm}\\
		\begin{minipage}{\mpColTwo}%
        \centering%
        \begin{tikzpicture}
        \draw (0, 0) node[inner sep=0] {\includegraphics[width=0.95\columnwidth]{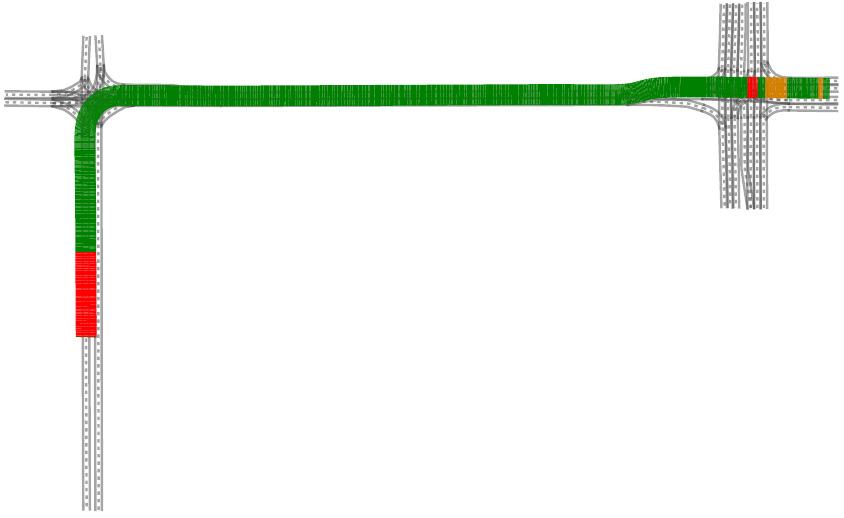}};
        \draw[text=black] (-3.8, -2) node {(d)};
        \end{tikzpicture}
		\end{minipage}
		\end{tabular}
	\end{minipage}
	\mpPostSpace
  \caption{\textbf{Rain scenario variation.} (a) Simulator, (b) Camera image, (c) World model (\apollo Dreamview), (d) output of \framework. 
  In adverse weather conditions the camera-based object detection wrongly detects an obstacle on the road floor, \framework detects the inconsistency and reports a high severity fault (red portion of the trajectory on the top-right). The red portion at the bottom-left corresponds to the high severity fault due to the overturned truck.}
	\end{center}
\end{figure}


\begin{figure}[ht!]
  \vspace{-1mm}
	\begin{center}
	\begin{minipage}{\textwidth}
	\begin{tabular}{c}%
		\begin{minipage}{\mpColTwo}%
        \centering%
        \begin{tikzpicture}
        \draw (0, 0) node[inner sep=0] {\includegraphics[width=0.95\columnwidth]{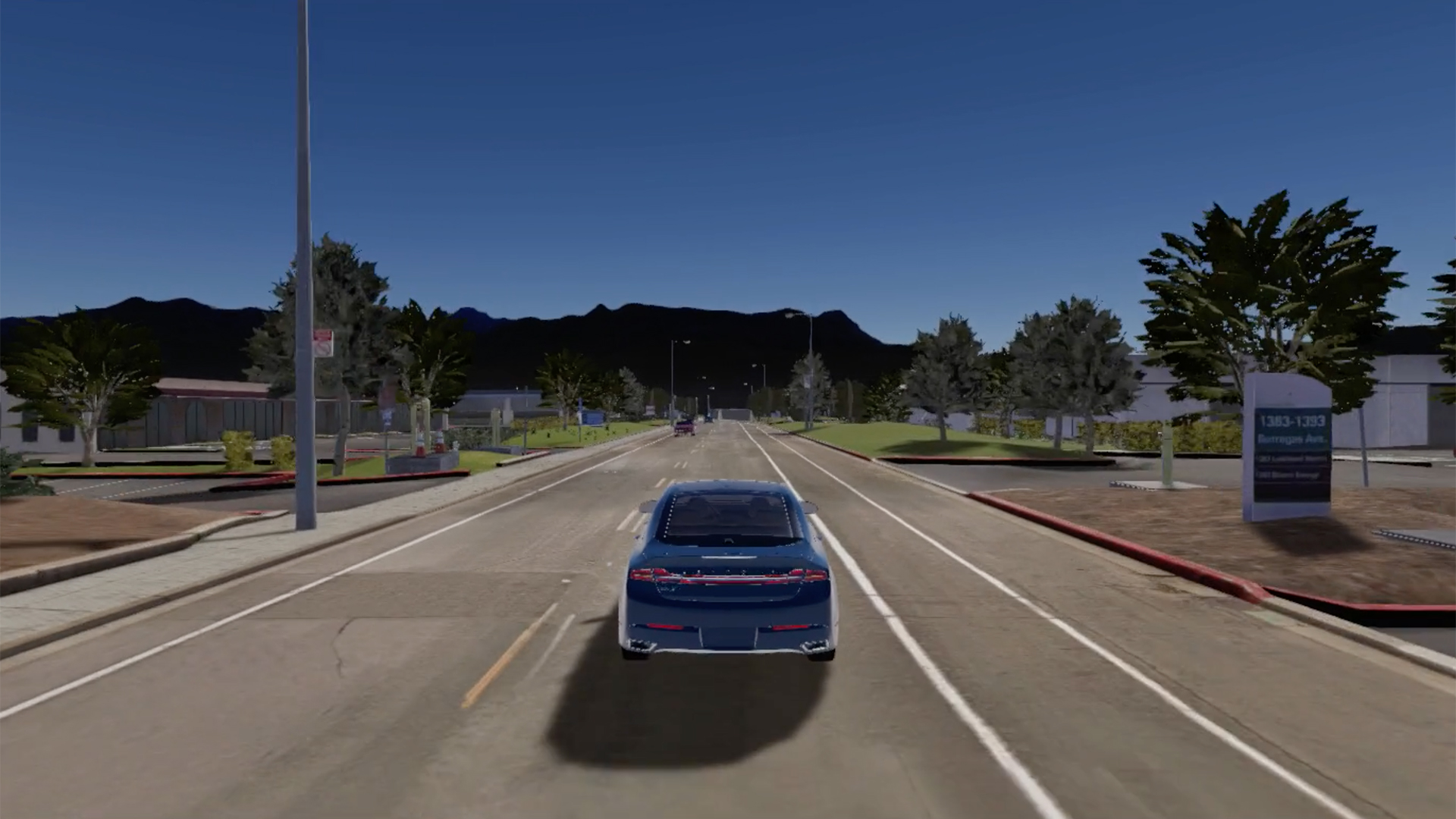}};
        \draw[text=white] (-3.8, -2) node {(a)};
        \end{tikzpicture}
		\end{minipage}
		\vspace{2mm}\\
		\begin{minipage}{\mpColTwo}%
        \centering%
        \begin{tikzpicture}
        \draw (0, 0) node[inner sep=0] {\includegraphics[width=0.95\columnwidth]{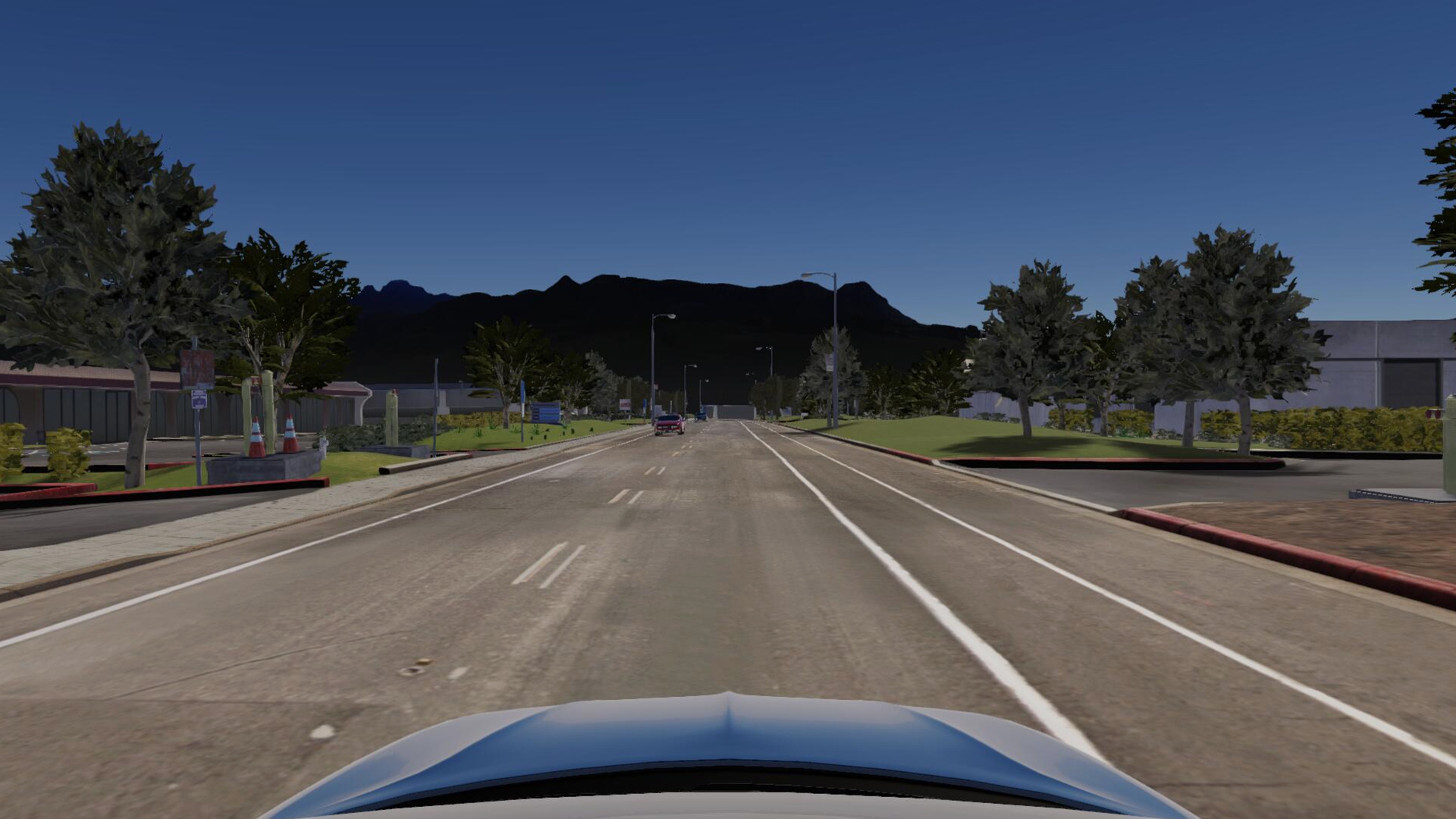}};
        \draw[text=white] (-3.8, -2) node {(b)};
        \end{tikzpicture}
		\end{minipage}
		\vspace{2mm}\\
		\begin{minipage}{\mpColTwo}%
        \centering%
        \begin{tikzpicture}
        \draw (0, 0) node[inner sep=0] {\includegraphics[width=0.95\columnwidth]{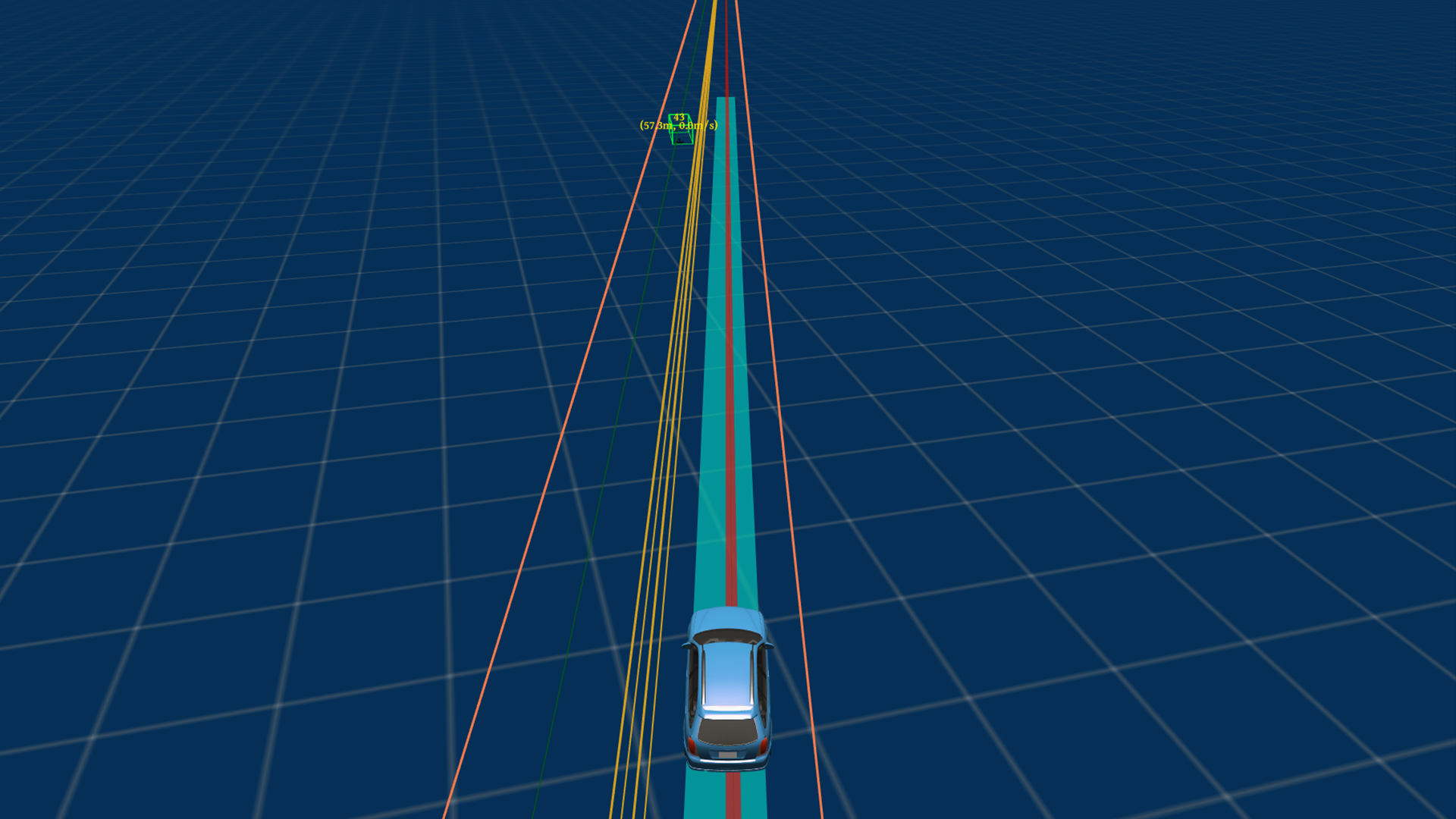}};
        \draw[text=white] (-3.8, -2) node {(c)};
        \end{tikzpicture}
    \end{minipage}
    \vspace{2mm}\\
		\begin{minipage}{\mpColTwo}%
        \centering%
        \begin{tikzpicture}
        \draw (0, 0) node[inner sep=0] {\includegraphics[width=0.95\columnwidth]{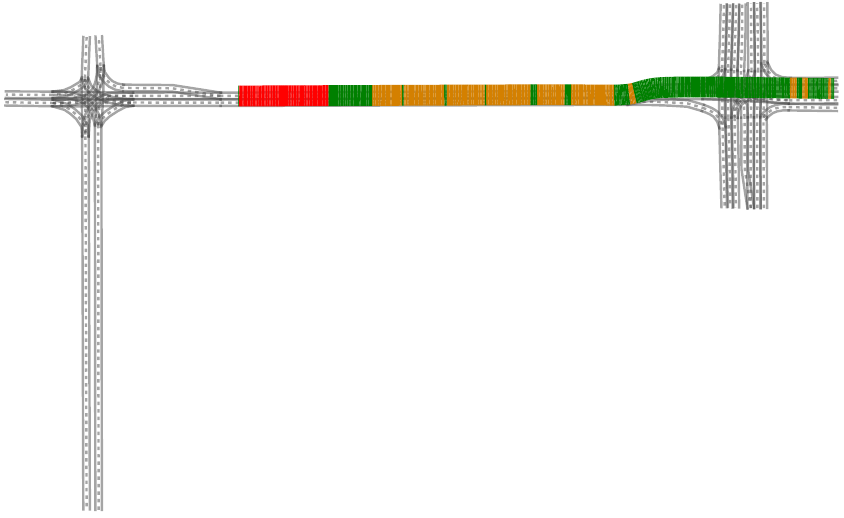}};
        \draw[text=black] (-3.8, -2) node {(d)};
        \end{tikzpicture}
		\end{minipage}
		\end{tabular}
	\end{minipage}
	\mpPostSpace
  \caption{\textbf{Traffic scenario variation.} (a) Simulator, (b) Camera image, (c) World model (\apollo Dreamview), (d) output of \framework.
  \apollo sensor fusion correctly detects the incoming car on the opposite lane but the camera fails to do so, as a result \framework reports low severity faults (yellow).
  In the final section of the trajectory \framework detects high severity faults (red) due to the overturned truck occupying the lane.}
	\end{center}
\end{figure}


\begin{figure}[ht!]
  \vspace{-1mm}
	\begin{center}
	\begin{minipage}{\textwidth}
	\begin{tabular}{c}%
		\begin{minipage}{\mpColTwo}%
        \centering%
        \begin{tikzpicture}
        \draw (0, 0) node[inner sep=0] {\includegraphics[width=0.95\columnwidth]{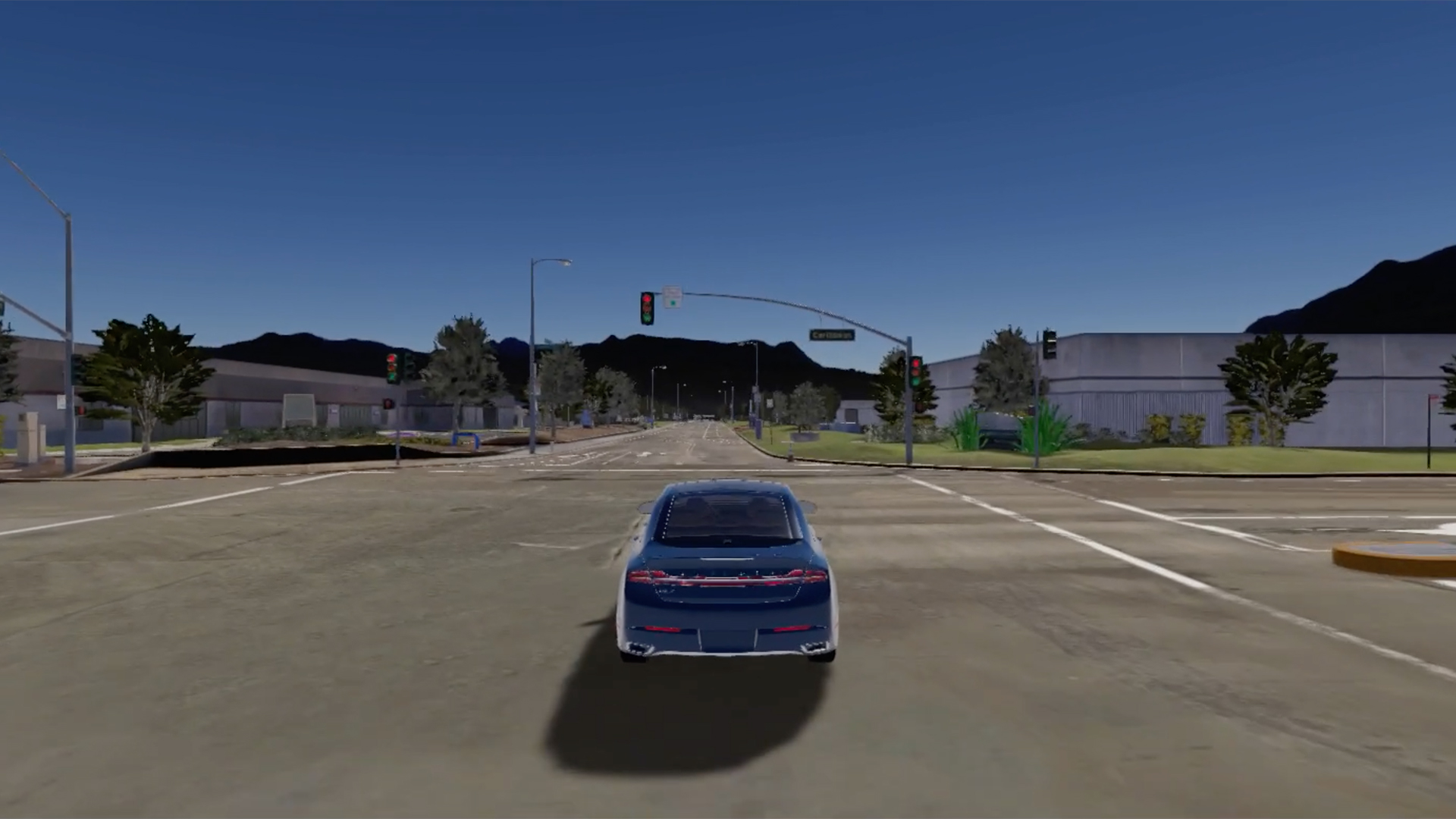} };
        \draw[text=white] (-3.8, -2) node {(a)};
        \end{tikzpicture}
		\end{minipage}
		\vspace{2mm}\\
		\begin{minipage}{\mpColTwo}%
        \centering%
        \begin{tikzpicture}
        \draw (0, 0) node[inner sep=0] {\includegraphics[width=0.95\columnwidth]{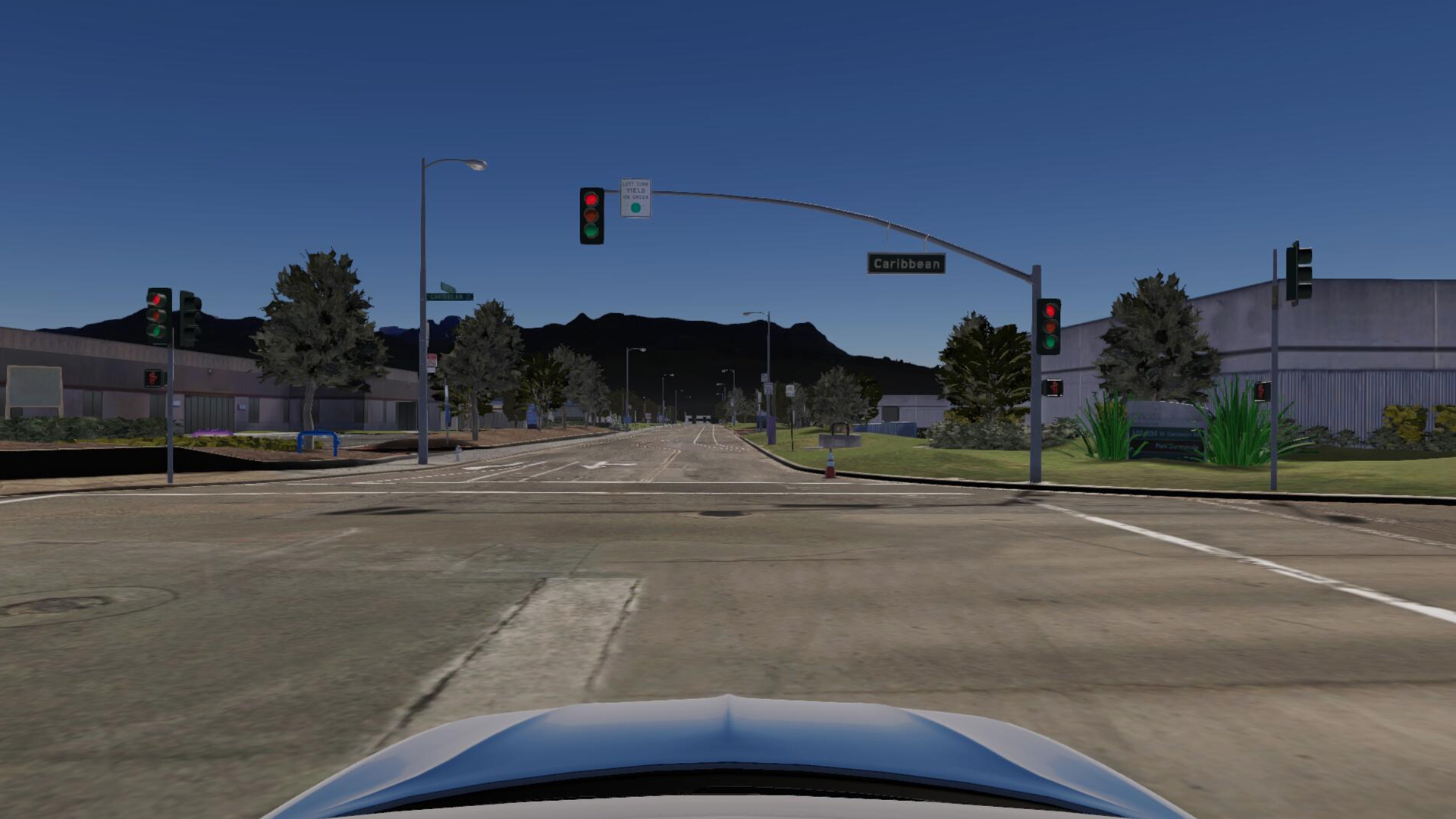}};
        \draw[text=white] (-3.8, -2) node {(b)};
        \end{tikzpicture}
		\end{minipage}
		\vspace{2mm}\\
		\begin{minipage}{\mpColTwo}%
        \centering%
        \begin{tikzpicture}
        \draw (0, 0) node[inner sep=0] {\includegraphics[width=0.95\columnwidth]{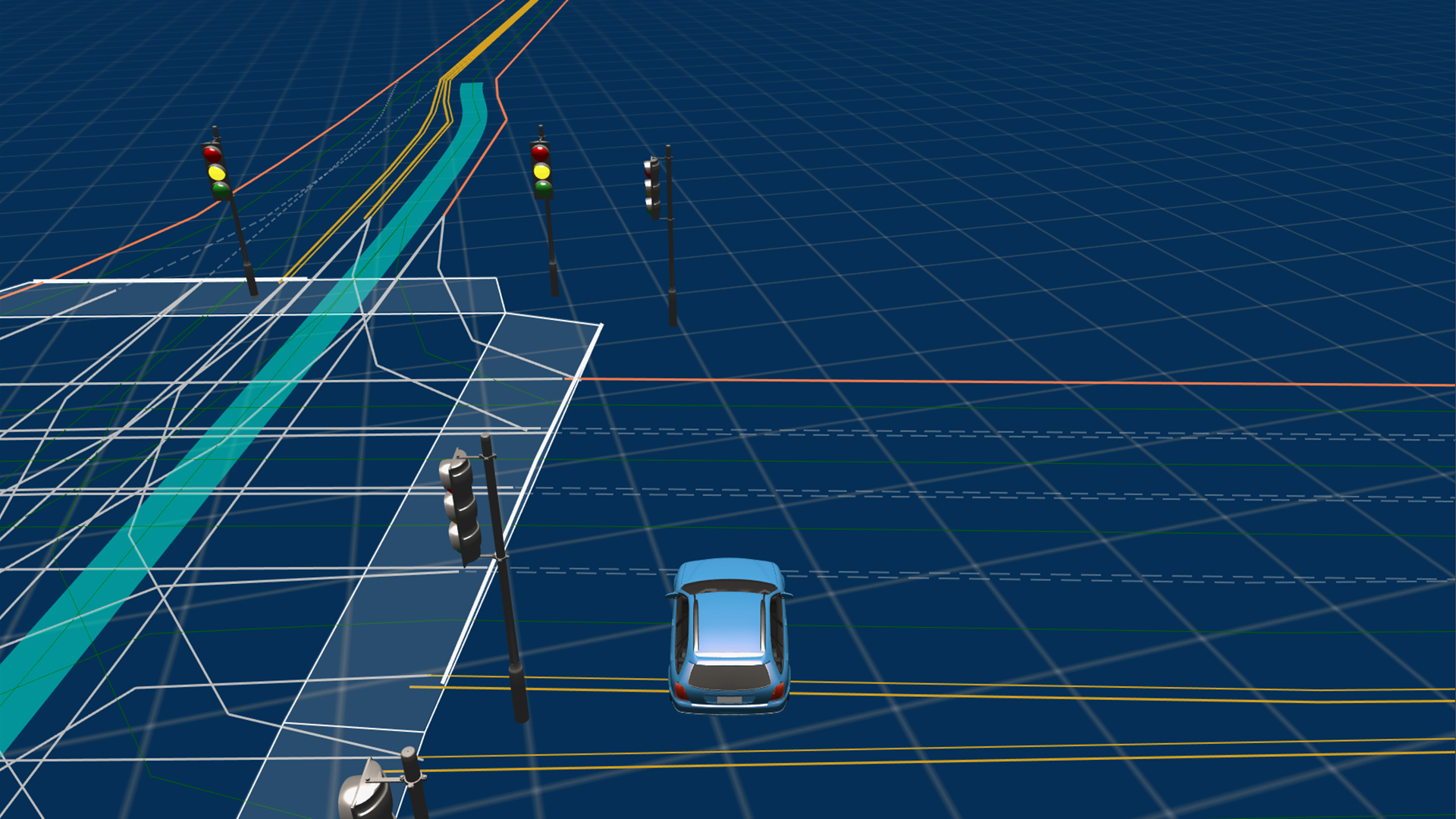}};
        \draw[text=white] (-3.8, -2) node {(c)};
        \end{tikzpicture}
    \end{minipage}
    \vspace{2mm}\\
		\begin{minipage}{\mpColTwo}%
        \centering%
        \begin{tikzpicture}
        \draw (0, 0) node[inner sep=0] {\includegraphics[width=0.95\columnwidth]{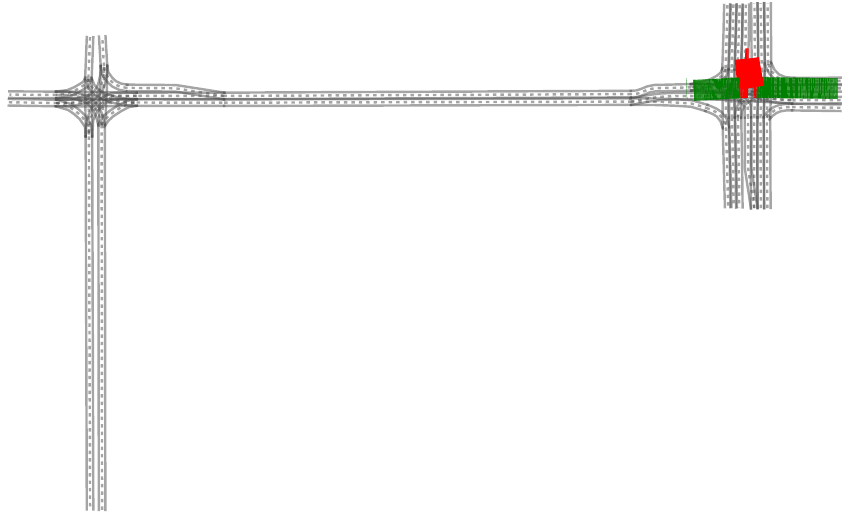}};
        \draw[text=black] (-3.8, -2) node {(d)};
        \end{tikzpicture}
		\end{minipage}
		\end{tabular}
	\end{minipage}
	\mpPostSpace
  \caption{\textbf{Spoiled GPS measurements.} (a) Simulator, (b) Camera image, (c) World model (\apollo Dreamview), (d) output of \framework.
  \apollo mislocalizes the AV due to spoiled GPS measurements, \framework detects inconsistencies between IMU and GPS and report a fault (red portion of the trajectory).}
	\end{center}
\end{figure}

}{
}

\end{document}